\documentclass[]{elsarticle}

\usepackage{lineno,xcolor}
\usepackage{amsmath,amssymb}
\usepackage{siunitx} 
\usepackage{hyperref}

\usepackage{enumitem}

\usepackage{wrapfig}
\usepackage{lscape}
\usepackage{rotating}
\usepackage{epstopdf}

\usepackage{tablefootnote}
\usepackage{soul}
\usepackage{subcaption} 
\usepackage{comment}
\usepackage{multirow}
\usepackage{booktabs}
\usepackage{comment}
\usepackage{xcolor}

\newcommand{\ie}{\textit{i.e.,}}

\newcommand\blfootnote[1]{%
  \begingroup
  \renewcommand\thefootnote{}\footnote{#1}%
  \addtocounter{footnote}{-1}%
  \endgroup
}

\begin{document}

\begin{frontmatter}

\title{GATraj: A Graph- and Attention-based Multi-Agent Trajectory Prediction Model}

\author[ITC]{Hao Cheng$^\dagger$}

\author[IKG]{Mengmeng Liu$^\dagger$\blfootnote{$\dagger$ Authors contributed equally}}

\author[VIS]{Lin Chen}

\author[VIS]{Hellward Broszio}

\author[IKG]{Monika Sester}

\author[ITC]{Michael Ying Yang}

\address[ITC]{Faculty of Geo-Information Science and Earth Observation (ITC), University of Twente, The Netherlands}

\address[IKG]{Institute of Cartography
and Geoinformatics, Leibniz University Hannover, Germany}

\address[VIS]{VISCODA GmbH, Germany}


\begin{abstract}
    Trajectory prediction has been a long-standing problem in intelligent systems like autonomous driving and robot navigation. 
    Models trained on large-scale benchmarks have made significant progress in improving prediction accuracy. 
    However, the importance on efficiency for real-time applications has been less emphasized. 
    This paper proposes an attention-based graph model, named \textit{GATraj}, which achieves a good balance of prediction accuracy and inference speed. 
    We use attention mechanisms to model the spatial-temporal dynamics of agents, such as pedestrians or vehicles, and a graph convolutional network to model their interactions. 
    Additionally, a Laplacian mixture decoder is implemented to mitigate mode collapse and generate diverse multimodal predictions for each agent. 
    GATraj achieves state-of-the-art prediction performance at a much higher speed when tested on the ETH/UCY datasets for pedestrian trajectories, and good performance at about \SI{100}{Hz} inference speed when tested on the nuScenes dataset for autonomous driving. 
    We conduct extensive experiments to analyze the probability estimation of the Laplacian mixture decoder and compare it with a Gaussian mixture decoder for predicting different multimodalities. 
    Furthermore, comprehensive ablation studies demonstrate the effectiveness of each proposed module in {GATraj}.    
    The code is released at \href{https://github.com/mengmengliu1998/GATraj}{GATraj}. 
    
\end{abstract}

\begin{keyword}
\texttt{ 
Trajectory prediction \sep 
Graph model\sep 
Autonomous driving\sep
Pedestrian\sep
Mixture density network
}
\end{keyword}
\end{frontmatter}

\section{Introduction}
\label{introduction}
Accurately predicting the movements of agents, such as pedestrians and vehicles, in various environments is crucial for many intelligent systems, including autonomous driving and robot navigation. 
One application is that, with accurate predictions of other agents' trajectories in the vicinity, an automated ego agent can safely navigate its own path. 
However, predicting trajectories is challenging for several reasons.
First, agents' behaviors are stochastic, as they mutually influence each other, for instance, by avoiding collisions or staying closely in a subgroup. 
Second, the information available to derive an agent's behavior is often limited, and their destination is typically unknown. 
In most cases, the ego agent can only estimate other agents' behaviors based on their perceived past moving dynamics, such as velocity and heading direction, and interactions depending on their relative positions with their environments.
Moreover, due to the mutual influence among agents and their movements in both spatial and temporal dimensions, an agent's behavior can be multimodal in terms of moving into different directions at various speeds. 
Therefore, trajectory prediction needs to consider both spatial-temporal dynamics and the multimodality of agents' behaviors.

\begin{figure}[t!]
\begin{center}
 \includegraphics[clip=true, trim=1.5cm 0.4cm 2cm 1cm, width=0.9\linewidth]{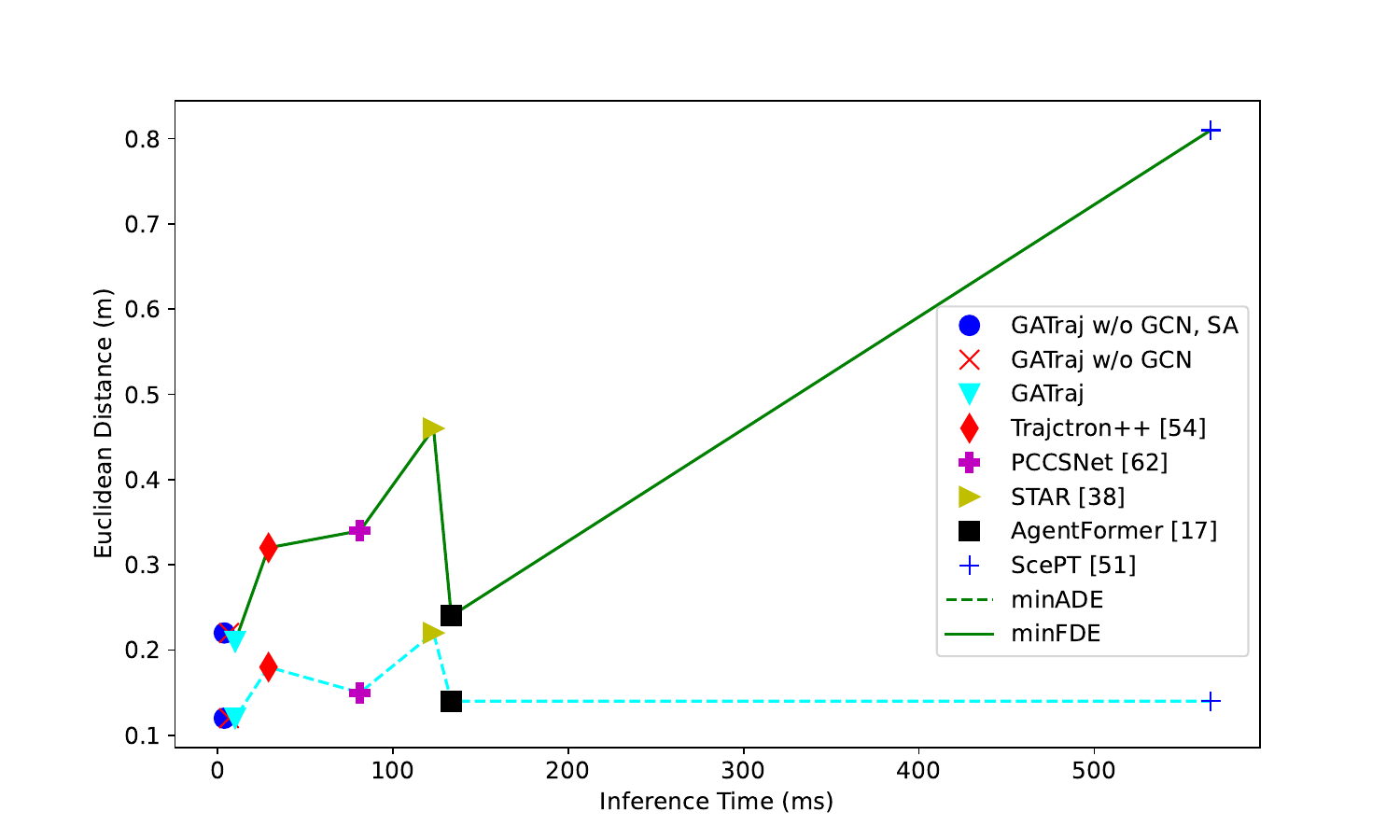}
\end{center}
   \caption{ The inference speed and the prediction performance of the
models on the Zara2 dataset \cite{lerner2007crowds} computed from a single Tesla V100 GPU. GATraj runs faster than other prediction methods and achieves superior performance. SA: self-attention \cite{vaswani2017attention}; GCN: graph convolutional network \cite{welling2017semi}.}
\label{fig:inference_time}
\end{figure}

The accuracy of trajectory prediction across multiple benchmarks has been significantly enhanced by recent deep learning models trained on large-scale real-world datasets, as demonstrated in Figure~\ref{fig:inference_time} by the y-axis representing the displacement errors measured by Euclidean distance. 
These models accomplish the prediction task by utilizing one or more of the following sources of information:

\noindent
\textbf{Observations of the past trajectories.}
The input for trajectory prediction tasks includes an agent's past trajectory in both lateral and longitudinal coordinates, as well as the trajectories of its neighboring agents observed in the vicinity. 
In pedestrian trajectory prediction tasks on the ETH/UCY benchmark \cite{pellegrini2009you,lerner2007crowds}, a popular time horizon setting is for the model to observe eight time steps and predict the next 12 time steps at a sampling rate of \SI{2.5}{Hz}. 
For autonomous driving on the nuScenes benchmark \cite{nuscenes}, the model typically observes up to four time steps and predicts the next 12 time steps at a sampling rate of \SI{2}{Hz}. 
More recent datasets, such as those introduced in \cite{chang2019argoverse,sun2020scalability,wilson2021argoverse}, pose even greater challenges with shorter observation periods and longer prediction horizons.

\noindent
\textbf{Interactions among agents.}
Modeling agent-to-agent interactions involves capturing their behavior over time and space. 
One popular technique is to use an occupancy map centered at the ego agent's position to map neighboring agents, known as the \textit{agent-centric} or marginal trajectory prediction approach \cite{wang2023prophnet}. 
This approach provides an invariant translation from a global coordination of all agents to the ego agent's local coordination, facilitating the interaction modeling using the occupancy map with a fixed ego perspective. 
Hence, this approach is widely used for data augmentation in trajectory prediction for autonomous vehicles \cite{varadarajan2022multipath++,nayakanti2022wayformer,gu2021densetnt}. 
However, inferring the positions of the neighboring agents through translation to the ego agent's local coordinate system can be time-consuming, as it requires iterating the translation one by one for each agent to be predicted.
Another approach is to predict the trajectories of all agents in the scene jointly, which is also called scene-centric approach \cite{sun2022m2i}.
This approach can speed up the inference process significantly but requires handling the variance in the global positions of each agent.
Alternatively, interactions among agents are modeled using graph models \cite{zhang2019sr}, in which agents are treated as nodes, their connections are modeled as edges, and the interaction information among them is conveyed via message passing.

\noindent
\textbf{Context constraints.}
Environmental scene contexts can be used to constrain an agent's movement. 
For example, convolutional and attention-based approaches can extract contextual information from rasterized data like raster maps and RGB images \cite{sadeghian2019sophie,phan2020covernet,yuan2021agentformer}, or vectorized data like High-Definition (HD) maps \cite{gao2020vectornet,gu2021densetnt}. 
However, maps may not be available or out dated for some areas of interest when a vehicle drives in.
Also, the scene context information may only have limited impact on pedestrians who walk freely in open environments or shared spaces.
To make prediction models applicable in any scene context settings, many models predict trajectories without using map information \cite{cheng2021amenet,yuan2021agentformer}.
In this work, we do not include any scene contextual information but focus more on the interactions among agents.

Primary criteria to evaluate a prediction model's performance are prediction errors between the predicted trajectories and the corresponding ground truth trajectories. 
Figure~\ref{fig:inference_time} shows that many recent studies on trajectory prediction prioritize reducing prediction errors, often competing to achieve centimeter-scale marginal improvement, while model efficiency (inference time) can vary from tens to several hundreds of milliseconds.
However, this high latency can limit the usage of complex trajectory prediction models in real-time scenarios, especially for autonomous driving systems that require millisecond-level response.
Therefore, in this paper, we aim to achieve a good balance of prediction performance and inference speed for multi-agent multimodal trajectory prediction.

Specifically, we propose {GATraj}, a multi-agent trajectory prediction model based on graph and attention mechanisms \cite{vaswani2017attention} that considers both prediction accuracy and inference speed. 
GATraj takes into account the spatial-temporal dynamics of agents and outputs multimodal trajectories. 
Attention mechanisms are used to capture agents' dynamics, while the graph convolutional network (GCN)-based module with message passing \cite{welling2017semi} is employed to model agent-to-agent interactions.
In order to speedup the prediction process, this interaction module is implemented in a scene-centric fashion, requiring no further time-consuming translation of the neighboring agent’s feature encodings to the ego agent’s local coordination.
Hence, unlike agent-centric models, GATraj can jointly predict all agents' trajectories in the given scene at a higher frequency.
Furthermore, we employ a Laplacian mixture decoder to predict diverse multimodal trajectories for each agent and a winner-takes-all training strategy to mitigate mode collapse.

The \textbf{main contributions} of our work are as follows:
\begin{itemize}
    \item We propose an end-to-end multimodal trajectory prediction model, named GATraj, which achieves a good balance of prediction performance and inference speed. It employs an efficient scene-centric GCN module to learn agent-to-agent interactions and an attention module to extract spatial-temporal dynamics.
    \item GATraj uses a Laplacian Mixture Density Network (MDN) decoder and a winner-takes-all \cite{makansi2019overcoming} training strategy, which produces more accurate probability estimation for the multimodal prediction than the widely used Gaussian MDN decoder. 
    \item GATraj achieves state-of-the-art prediction performance with significantly higher prediction speed, as demonstrated by testing on the ETH/UCY benchmark datasets for pedestrian trajectories. Additionally, it achieves on-par prediction performance at approximately \SI{100}{Hz} for real-time inference on nuScenes for autonomous driving.
\end{itemize}

\section{Related Work}
\label{relatedwork}

In this section, we discuss the development of trajectory prediction for pedestrians and vehicles based on the methods applied for modeling sequential dynamics, interactions, and multimodalities of trajectories.

\subsection{Modeling Motion Dynamics as a Time Sequence}
The transition of an agent's motion dynamics, namely the change in speed profile, can be simplified as a temporal sequence of states. 
Methods such as Kalman Filter \cite{kalman1960new}, Gaussian Process \cite{kim2011gaussian}, and Markov Models \cite{kitani2012activity} have been commonly used for trajectory prediction. However, these techniques have limited performance when it comes to cope with increased temporal complexity.

In recent years, data-driven models with an encoder-decoder structure have become the dominant approach to trajectory prediction modeling. 
Specifically, Recurrent Neural Networks (RNNs) \cite{rumelhart1986learning} and their variants Long Short-Term Memories (LSTMs) \cite{hochreiter1997long} and Gated Recurrent Units (GRUs) \cite{cho2014properties} have been used to gate the information for updating the states in a sequence. 
Additionally, attention mechanisms \cite{vaswani2017attention}, widely used in Natural Language Processing (NLP), have shown their effectiveness in learning complex spatial-temporal interconnections and have been adopted in trajectory prediction. 
The attention mechanisms guide the interconnections between states, helping to address long time-dependency and complex connectivity problems. 
These models have achieved state-of-the-art performance on various trajectory prediction benchmarks \cite{liu2021multimodal,gu2022stochastic,zhou2022hivt,ngiam2022scene}.
Alternatively, the history trajectories are stored and later retrieved to identify similar motion dynamics. 
This history information is later treated as reference in the memory-based model SHENet \cite{meng2022forecasting} and instance-based model MemoNet \cite{xu2022remember} to guide the predictions in the future.  

\subsection{Modeling Interactions Among Agents}
Early works in trajectory prediction often relied on hand-crafted features to model interactions among agents.
One of the most influential methods is the Social Force Model (SFM) \cite{helbing1995social}, which applies different forces to determine agents' speed and orientation. 
These forces include a repulsive force for collision avoidance with obstacles and an attractive force for social connections among agents and goals. 
Game theoretic models \cite{johora2022generalizability} have also been used to simulate the negotiations of right-of-way among agent players for decision making. 
However, due to the complex dynamics in both spatial and temporal domains, these models based on manually selected or designed features often have limited performance in modeling multi-agent interactions and the multimodalities of potential future trajectories.

In recent years, interactions among agents have been modeled by aggregating latent variables learned from each agent's motion dynamics. 
The pioneering work Social-LSTM \cite{alahi2016social} explores LSTMs \cite{hochreiter1997long} to encode pedestrians' motion dynamics into hidden states and a pooling mechanism to model interactions. 
Many later works \cite{xue2018ss,cheng2021amenet} have extended this structure by including more features, such as agent-to-agent and agent-to-environment interactions using LSTMs or GRUs \cite{cho2014properties}. However, the model performances heavily depend on the hidden states, and with the increase in trajectory length and complexity, the performances are often significantly degraded \cite{Hug2021Quantifying}.

Attention-based \cite{vaswani2017attention} and graph models, such as Graph Convolutional Networks (GCNs) \cite{welling2017semi}, have been leveraged to model agent-to-agent interactions in trajectory prediction.
The self- and cross-attention mechanisms have shown their effectiveness in learning interaction information \cite{yu2020spatio,yuan2021agentformer}.
For instance, AgentFormer \cite{yuan2021agentformer} proposes an agent-aware attention mechanism that simultaneously models the temporal and social dimensions among agents.
In addition to attention mechanisms, GCNs are widely used to model interactions in trajectory prediction \cite{zhang2019sr,shi2020multimodal,gilles2022gohome,bae2022learning}.
Agents are represented as nodes, and their connections are represented as edges, allowing for message passing between nodes to capture interactions between agents.
This paper describes an implementation of attention mechanisms to learn salient spatial-temporal features and a GCN module to model agent-to-agent interactions.

\subsection{Multimodal Prediction}
Multimodal prediction refers to the task of predicting a set of feasible trajectories for each agent, accounting for their stochastic behavior, such as varying speeds and directions. 
Deep generative models are commonly used to address the multimodality problem, including Generative Adversarial Nets (GANs) \cite{goodfellow2014generative}, Variational Auto-Encoder (VAE) \cite{kingma2014auto} and its extension Conditional-VAE (CVAE) \cite{kingma2014semi}, and Normalizing Flows \cite{rezende2015variational}. Social GAN \cite{gupta2018social}, DESIRE \cite{lee2017desire}, and Precog \cite{rhinehart2019precog} are early representative trajectory prediction frameworks that apply these designs, respectively.
Recent works such as \cite{cheng2021exploring,yuan2021agentformer,lee2022muse,chen2022scept} have extended the CVAE-based design for multimodal trajectory prediction due to its good performance.
However, these sampling-based approaches do not provide a straightforward mechanism to estimate the likelihood of each prediction in the random sampling process.
In addition, with the success in computer vision domains, diffusion models \cite{ho2020denoising} have been adopted to learn road users' behavior and generate diverse multimodal trajectory predictions \cite{gu2022stochastic,mao2023leapfrog}. However, these models are often time consuming due to the chain of sampling process.

Apart from deep generative models, MDNs are proposed to learn a mixture density function, such as Gaussian Mixture Model (GMM), for multimodal trajectory prediction \cite{salzmann2020trajectron++,shi2020multimodal,varadarajan2022multipath++,deo2022multimodal}. 
 However, most benchmarks only provide a single ground truth trajectory, making it difficult for these models to learn the entire data distribution and generate diverse predictions. 
 This issue, referred to as the mode collapse problem \cite{richardson2018gans}, has been a challenge for these methods.
The winner-takes-all strategy is proposed to mitigate the mode-collapse problem \cite{makansi2019overcoming,zhou2022hivt,deo2022multimodal}, which only optimizes the loss function of the best predicted modality to facilitate the training process and encourage more diverse predictions.

Inspired by these previous methods, we design our model, GATraj, which uses an encoder-decoder structure to encode past trajectories and interactions, and extract spatial-temporal features through attention mechanisms. 
In addition, we utilize a GCN to model interactions among agents. 
Different from the previous models, we seek for a more efficient framework with a scene-centric GCN module to learn global interactions among agents and jointly predict multimodal trajectories of each agent in the given scene simultaneously. 
This scene-centric GCN module avoids the time-consuming translation into the ego agent's local coordination, and significantly speed up the inference process.
Rather than relying on a GMM, we introduce a Laplacian Mixture Model (LMM) as the decoder with the winner-takes-all training strategy that generates multiple diverse predictions for each agent.

\begin{figure*}[hbpt!]
\begin{center}
 \includegraphics[clip=true, trim=0in 5.5in 3.3in 0in, width=\linewidth]{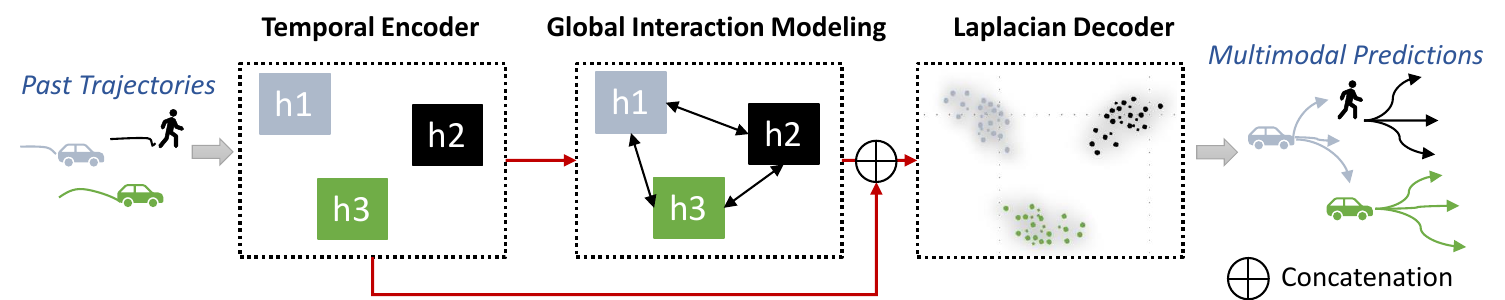}
\end{center}
   \caption{The framework of the proposed model GATraj, consisting of three parts: a Temporal Encoder, a GCN-based global interaction module, and a Laplacian Decoder. It takes as input the observed trajectory of each agent and outputs multimodal predictions of their potential future trajectories.  
}
\label{fig:framework}
\end{figure*}

\section{METHODOLOGY}
\label{methodology}
\subsection{Problem Formulation}
\label{problemformulation}
A multi-path trajectory prediction problem is defined as predicting a set of future trajectories $\{\hat{\mathbf{Y}}^{T+1:T+T'}_{i,1},\dots,\hat{\mathbf{Y}}^{T+1:T+T'}_{i,K}\}$ given the observed trajectory $\mathbf{X}^{1:T}_{i}=\{X^{1}_{i},...,X^{T}_{i}\}$ of agent $i$.  
Here, $T$ and $T'$ denote the total time steps of the observed and the predicted trajectories, respectively; $T+T'$ is the total sequence length.
$K$ stands for the number of the modes of multiple predicted trajectories.
$X^{t}_{i}=\{x^{t}_{i},y^{t}_{i}\}\in \mathbb{R}^2$ is the position of agent $i$ at time step $t$ in a 2D coordinate system. 
The formulation can also be easily extended to a 3D coordinate system.
To simplify the notation, $\mathbf{X}$ and $\mathbf{Y}$ denote the observed and ground truth trajectories, respectively.
The loss function aims to reduce the distance between a predicted trajectory $\hat{\mathbf{Y}}$ and the corresponding ground truth trajectory $\mathbf{Y}$ for all agents.

The input trajectories are shifted before feeding the trajectory data into the prediction model. 
Specifically, we follow \cite{cheng2021exploring} to use relative positions instead of absolute positions to mitigate domain gaps across different scenes. 
First, we shift the origin to each agent's last observed time step for data normalization.
Then, we use $\Delta \mathbf{X}^{2:T}_{i}=\{\Delta X^{2}_{i},...,\Delta X^{T}_{i}\}$ to represent agent \textit{i}'s observed trajectory, where $\Delta X^{t}_{i}=\{\Delta x^{t}_{i},\Delta y^{t}_{i}\}$ is the offset from $(t-1)$ to the next time step.
However, this shifted representation of trajectories also loses the relative position information between agents, which is essential for modeling their interactions. 
Hence, we obtain the relative position between agent $i$ and $j$ based on their original positions $(x^{t}_{i}-x^{t}_{j}, y^{t}_{i}-y^{t}_{i})$ at each time step, where $i \neq j$. The relative position is further employed in the global interaction modeling.

\subsection{The Proposed Framework}
\label{subsec:framework}
Figure~\ref{fig:framework} depicts the overview of our proposed framework GATraj. 
It consists of three parts: a Temporal Encoder, a GCN-based global interaction module, and a Laplacian Decoder. 
We first utilize the encoder to extract temporal information of each agent independently and output rich temporal information for the subsequent modules. 
Then the global interaction module, which employs a GCN adopted from  \cite{zhang2019sr}, aggregates the temporal context of different agents over time and space and updates each agent’s hidden state by message passing for interaction modeling. 
Finally, the learned spatial-temporal feature map is the input of the Laplacian decoder that simultaneously predicts diverse and multimodal future trajectories for all the agents. 
We explain each part of GATraj in detail in the following.

\noindent
\paragraph{\textbf{Temporal Encoder}}
\label{temporalencoder}
\begin{figure*}[hbpt!]
\begin{center}
 \includegraphics[clip=true, trim=-2in 5.4in 6.5in 0in, width=\linewidth]{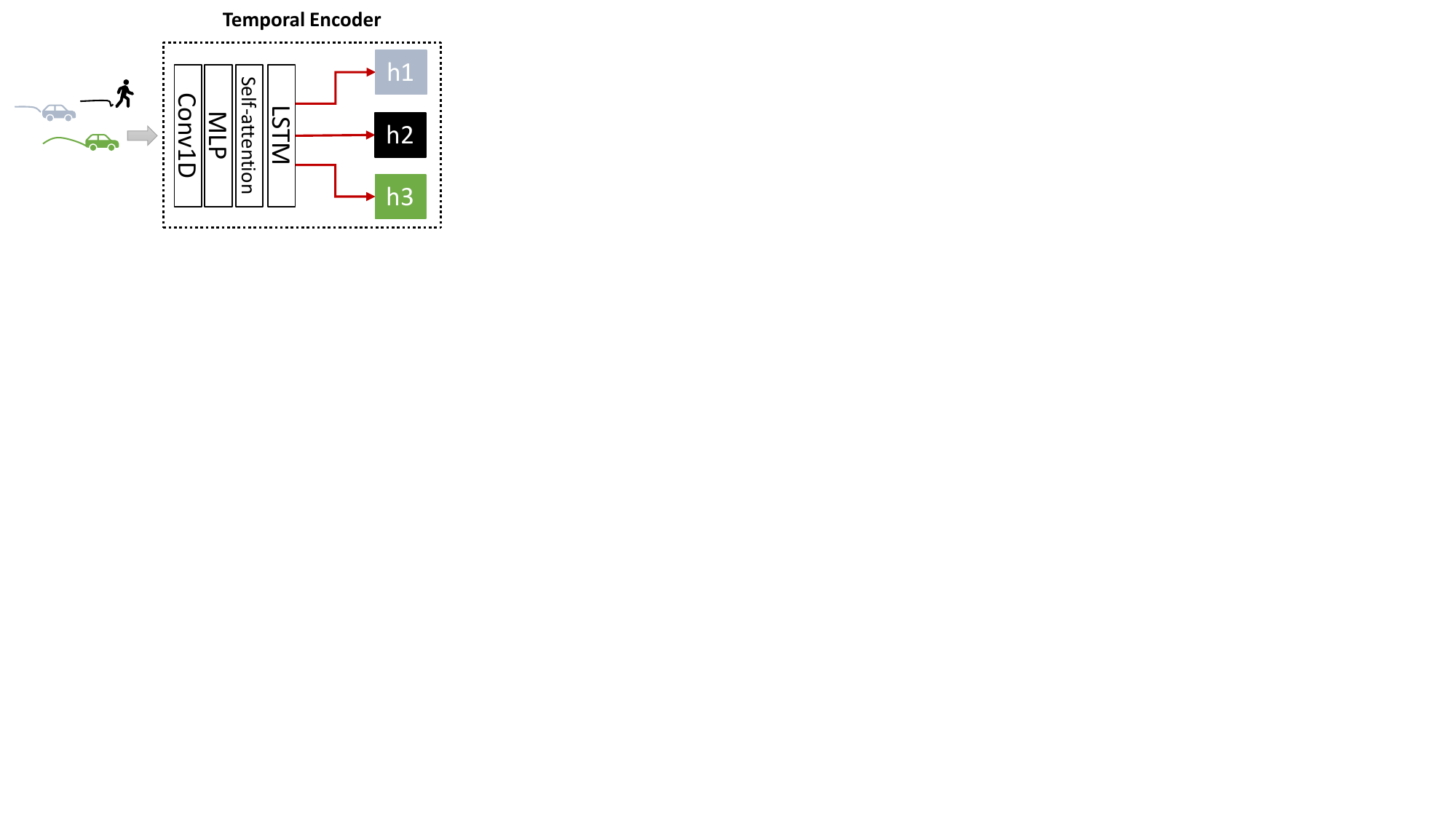}
\end{center}
   \caption{The Temporal Encoder of GATraj. It consists of a sequence of 1D convolution layer (Conv1D) and a two-layer position-wise multilayer perceptron (MLP), a self-attention layer, and an LSTM layer to learn salient temporal features of each agent.
}
\label{fig:framework-encoder}
\end{figure*}

The self-attention Temporal Encoder learns temporal dynamic information from the observed trajectory.
Concretely, as shown in Figure \ref{fig:framework-encoder}, the temporal encoder takes as input the relative positions $\Delta X^{2:T}_{i}$, which is later passed to a 1D convolution layer (Conv1D) and a two-layer position-wise multilayer perceptron (MLP). 
In the following, the self-attention layer with three Transformer Encoder blocks \cite{vaswani2017attention} learns salient temporal features - to which time steps of the temporal sequence the encoder should pay more attention through the self-attention layer.
Specifically, the default positional encodings of the Transformer network are added at the bottom of the encoder blocks to retain the ordered sequential information.
After that, a multi-head of self-attention is applied to jointly attend to the information from different representation subspaces at different positions.
Eight heads of self-attention and skip-connections are used in our implementation.
Towards the end of the temporal encoder, we utilize an LSTM to extract the temporal dependencies over time. 
It should be noted that each agent's observed trajectory is considered independently so that their temporal dynamics are processed in parallel.

\noindent
\paragraph{\textbf{Global Interaction}}
We build a GCN-based module, inspired by SR-LSTM \cite{zhang2019sr}, for modeling global interactions among all the concurrent agents in a given scene.
SR-LSTM introduces a States Refinement (SR) module to refine the cell state of an LSTM by passing messages among agents at each time step.
In contrast, to reduce the computational cost, GATraj only refines cell state once at the latest observed time step but still maintains the ability to model interactions among agents.
More specifically, a motion gate and an agent-wise attention \cite{zhang2019sr} are adopted to preserve spatial relationships between the ego agent and its neighbors and to select the most helpful information from neighboring agents for message passing.
The cell state output by the LSTM in the Temporal Encoder is updated by a social-aware information selection mechanism as follows:
\begin{equation}
\label{eq:socialinfor}
\hat{c}^{T,\,l+1}_{i} = \phi_{\text{mp}}[\sum_{j \in \mathsf{N}(i)} \alpha^{T,\,l}_{i,\,j}(g^{T,\,l}_{i,\,j}\odot \hat{h}^{T,\,l}_{j})] + \hat{c}^{T,\,l}_{i},
\end{equation}
where $\hat{c}_{i}$ denotes agent $i$'s cell state after message passing, $l$ denotes the message passing times, $\odot$ is the element-wise product operation, and $\phi_{\text{mp}}$ denotes an MLP. 
Namely, for the ego agent $i$, the cell state starts at $l = 0$ with the original output of the LSTM ${c}_{i}$. 
All the LSTM hidden states $\hat{h}^{T,l}_{j}$ of its neighbors $j \in \mathsf{N}(i)$ are aggregated through the motion gate $g^{T,l}_{i,j}$, and the agent-wise attention $\alpha_{i,j}$ attends to the neighbors based on the agent-to-agent pairwise weights. 
In addition, a skip connection adds agent $i$'s previously refined cell $\hat{c}^{T,l}_{i}$. 
Here, the motion gate is defined by using the relative position between agent $i$ and $j$, and their individual hidden states, as shown in Eq.~\eqref{eq:relative_pos} and \eqref{eq:motion_gate},
\begin{equation}
\label{eq:relative_pos}
r^{T,\,l}_{i,j} =\phi_{r}(x^{T}_{i}-x^{T}_{j},y^{T}_{i}-y^{T}_{j}),
\end{equation}
\begin{equation}
\label{eq:motion_gate}
g^{T,l}_{i,j} =\delta(\phi_{m}[r^{T,l}_{i,j}, \hat{h}^{T,l}_{j}, \hat{h}^{T,l}_{i}]),
\end{equation}
where $\phi_{m}$ and $\phi_{r}$ denote MLP and $\delta$ is the Sigmoid function.

Similarly, another MLP $\phi_{a}$ is used to learn the weights $u^{T,\,l}_{i,\,j}$ of the different impacts from the neighbors by Eq.~\eqref{eq:weight}. The weights are normalized across all the neighboring agents using the Softmax function denoted by Eq.~\eqref{eq:normalize},
\begin{equation}
\label{eq:weight}
u^{T,\,l}_{i,\,j} = \phi_{a}[r^{T,\,l}_{i,\,j},\, \hat{h}^{T,\,l}_{j},\, \hat{h}^{T,\,l}_{i}],
\end{equation}
\begin{equation}
\label{eq:normalize}
\alpha^{T,\,l}_{i,\,j} = \frac{\exp(u^{T,\,l}_{i,\,j})}{\sum_{s \in \mathsf{N}(i)} \exp(u^{T,\,l}_{i,\,s})}.
\end{equation}

The hidden state of agent $i$ then is updated using Eq.~\eqref{eq:update_hidden} after the cell state $\hat{c}^{T,\,l+1}_{i}$ is refined by the above message passing.
\begin{equation}
\label{eq:update_hidden}
\hat{h}^{T,\,l+1}_{i} = \hat{h}^{T,\,l}_{i} + \text{tanh}(\hat{c}^{T,\,l+1}_{i}),
\end{equation}
where tanh stands for the hyperbolic tangent function.
Different from SR-LSTM, we use a skip-connection to avoid the vanishing gradient problem and discard the output gate for a lightweight structure.
Compared to the original GCNs \cite{welling2017semi} that use an adjacency matrix to compute the normalization constant, our agent-wise attention uses learnable MLP $\phi_{r}$ and $\phi_{a}$ to aggregate relative spatial positional information between the ego and all its neighboring agents and emphasize important neighbors using the attention mechanism to guide the message passing.
Moreover, as the interactions among agents are automatically learned by using the above message passing and the agent-wise attention mechanism, there is no further need to translate the neighboring agent's feature encodings to the ego agent's local coordination.
This significantly reduces the processing time.

\noindent
\paragraph{\textbf{Laplacian Decoder}}
\label{laplacedecoder}
\label{globalinteraction}
\begin{figure*}[hbpt!]
\begin{center}
 \includegraphics[clip=true, trim=-0.5in 5.1in 5in 0in, width=\linewidth]{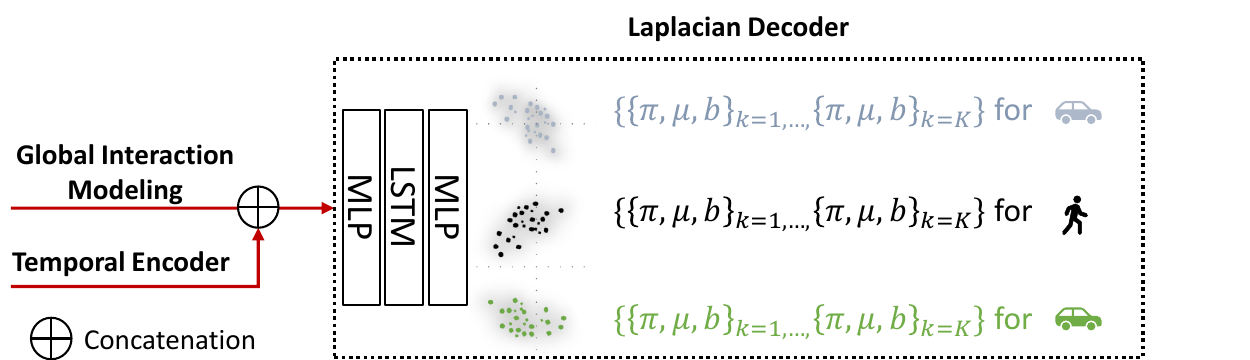}
\end{center}
   \caption{The Laplacian Decoder of GATraj, which consists of a sequence of MLP, LSTM and MLP layers. It takes as input the spatial-temporal dynamic context from the temporal encoder and the GCN module of the global interaction modeling and then outputs a set of parameters, i.e., the likelihood of the mode $\pi$, and the location $\mu$ and scale $b$ of a Laplacian mixture model for each agent. 
}
\label{fig:framework-decoder}
\end{figure*}
We introduce a Laplacian MDN decoder to generate future multimodal trajectories, accounting for each agent's stochastic behavior.
The decoder takes as input the spatial-temporal dynamic context from the temporal encoder and the GCN module of the global interaction modeling.
Its outputs are a set of different modes of the predicted trajectory distribution. 
Concretely, the predicted set of potential trajectories is 
$\{\{\pi, \mu, b\}_{k=1}, ..., \{\pi, \mu, b\}_{k=K}\}$ with a total of $K$ modes. For mode $k$, $\pi$ stands for the estimated likelihood of this mode among all the potential modes and $\sum^{K}_{k=1}{\pi}_{k}=1$.
In the Laplace distribution of the predicted mode, $\{\mu, b\} \in \mathbb{R}^{t_f\times 2}$, the location $\mu$ and scale $b$ parameters represent the mean positions and standard deviations of the predicted trajectory, respectively. 
To simply the generation of each trajectory, in this paper, we use the sequence of the mean positions $\mu$ as the predicted trajectory.

To be more specific, in the decoding process, the decoder takes as input the hidden state $h^{T}_{i}$ of the LSTM from the Temporal Encoder and  
the hidden state $\hat{h}^{T,\,l+1}_{i}$ and cell state $\hat{c}^{T,\,l+1}_{i}$ after the message passing with the GCN module.
$\{ h^{T}_{i}, \, \hat{h}^{T,\,l+1}_{i}, \, \hat{c}^{T,\,l+1}_{i}\} \in \mathbb{R}^{N \times D} $,
where $D$ is the dimension of the embedded feature space and $N$ is the total number of agents in the current scene. 
First, an MLP projects the shape of the input into $[K,\, N,\, D]$, where $K$ is the number of modes to be predicted. With projected feature embeddings, we utilize an MLP and a Softmax function to learn the probability $\hat{\pi}_{i,\,1:K}$ of each mode for each agent.
Then, an LSTM decodes the aggregated and embedded hidden states into a shape of $[K \times N,\, T',\, D]$, recovering the prediction time-step dimension $T'$. 
It should be noted that here we simultaneously predict all the time steps using the hidden states instead of applying a step-wise auto-regressing to further speed up the prediction. 
We empirically found that using LSTM contributes to a more efficient gating of the sequential information over the time axis (more details in Sec.~\ref{ablationstudies}).
Finally, two side-by-side MLPs predicts a mixture of Laplace distribution with $K$ modes of the potential future trajectories for each agent, \ie~the location $\hat{\mathbf{Y}}_{i,\,1:K} \in \mathbb{R}^{K\times T' \times 2}$ and its associated scale $\hat{\textbf{b}}_{i,\,1:K} \in \mathbb{R}^{K\times T' \times 2}$,
and $\hat{\pi}_{i,1:K} \in \mathbb{R}^{K}$.

\subsection{Loss Function}
\label{training}
The total loss of GATraj is decoupled into two parts - \textit{regression loss} and \textit{classification loss}. 
We utilize a Winner-Takes-All strategy \cite{makansi2019overcoming} for the supervision by each agent's single ground truth trajectory to encourage GATraj to generate diverse predictions. 
Namely, for the regression loss, we only optimize the best mode $k^*$ of the $K$ predictions during training instead of the weighted strategy by an Expectation-Maximization algorithm for a GMM. 
Following \cite{zhou2022hivt}, we employ the negative log-likelihood for the Laplace distribution as the regression loss and the cross-entropy loss as the classification loss for the mode optimization,
\begin{equation}
\label{eq:winner}
k^* =  \underset{{k\in K}}{\text{arg\,min}}||\hat{\mathbf{Y}}_{i,k} - \mathbf{Y}_{i}||^{2},  
\end{equation}
\begin{equation}
\label{eq:regressionloss}
\mathcal{L}_{\text{reg},\,i}=\frac{1}{T'}\sum^{T+T'}_{t=T+1}-\log P(\mathbf{Y}^{t}_{i}|\hat{\mathbf{Y}}^{t}_{i,\,k^*}, \mathbf{b}^{t}_{i,\,k^*}),
\end{equation}
\begin{equation}
\label{eq:clsloss}
\mathcal{L}_{\text{cls},\,i}=\sum^{K}_{k=1}-\pi_{i,\,k}\log(\hat{\pi}_{i,\,k}),
\end{equation}
where $\log P(\cdot \mid \cdot)$ is the logarithmic probability density function of the Laplace distribution. $\hat{\pi}_{i,k}$ is the predicted probability, and $\pi_{i,k}$ is our target probability.
We employ a soft displacement error adopted from \cite{zhou2022hivt} as our target probability.

\section{Experiments}
\label{experiment}

\subsection{Datasets}
\label{dataset}
The benchmarks ETH/UCY \cite{pellegrini2009you,lerner2007crowds} and nuScenes \cite{nuscenes} are used to train and test the performance of our proposed model.
ETH/UCY contains thousands of pedestrian trajectories sampled in \SI{2.5}{Hz} in five different datasets, \ie~Eth, Hotel, Uni, Zara1, and Zara2. 
The time horizon lets the model observe eight steps and predict the subsequent 12-step trajectories.
We follow the standard leave-one-out training and test partitioning, in order to test the model's generalization performance in unseen scenes. 
Concretely, four of the five datasets are used for training, and the left-out one is for the test.
This partitioning is iterated for each dataset.
The nuScenes prediction task provides separate training and test sets for vehicle trajectory prediction.
In total, nuScenes includes 1000 driving scenarios in different cities. Each scenario contains various neighboring agents and the target vehicles for the motion prediction task. 
In this dataset, a trajectory is sampled in \SI{2}{Hz}.
The model observes up to four-time steps and predicts the following 12-step trajectories.
We use the provided data partitioning, namely, 750 scenarios for training, 150 scenarios for validation, and the remaining 150 scenarios for testing.

\subsection{Evaluation Metrics}
\label{evaluation}
We apply the two most commonly used metrics, Average Displacement Error (ADE) and Final Displacement Error (FDE) in meters, to evaluate trajectory prediction accuracy. 
ADE measures the Euclidean distance between the predicted and ground truth trajectories and is averaged at each position for each agent.
FDE is the Euclidean distance of the last position between the predicted and ground truth trajectories.
\begin{equation}
\label{eq:ADEK}
\text{ADE}_{K}=\frac{1}{N}\frac{1}{T'}\text{min}^{K}_{k=1}\sum^{N}_{i=1}\sum^{T+T'}_{t=T+1}||\hat{y}^{t}_{i,\,k}-y^{t}_{i}||^{2},
\end{equation}
\begin{equation}
\label{eq:FDEK}
\text{FDE}_{K}=\frac{1}{N}\text{min}^{K}_{k=1}\sum^{N}_{i=1}||\hat{y}^{T+T'}_{i,\,k}-y^{T+T'}_{i}||^{2},
\end{equation}
where $N$ is the total number of agents.
Here, $K$ denotes that we generate $K$ predictions for each agent and report the best one measured by ADE and FDE, respectively.

Moreover, to evaluate the performance of the predicted distribution of a mixture model, we use negative log-likelihood (NLL) to compare the likelihood estimation across different models tested on the same data.
We report the average NLL and standard deviation across all the agents.

\subsection{Baselines}
\label{baselines}

To make a fair comparison, we only benchmark our model GATraj with multi-path trajectory prediction models.
We compare the GATraj's performance with the following models on the nuScenes dataset for autonomous driving.
\begin{itemize}
    \item MTP ({\color{blue}ICRA'19}) \cite{cui2019multimodal} is a CNN-based method that encodes each agent's surrounding context into a raster image and predicts a set of possible trajectories. 
    \item MultiPath ({\color{blue}CoRL'20}) \cite{chai2019multipath} uses a set of anchors to regress the predicted trajectory distribution, resulting in a Gaussian mixture at each time step. 
    \item DLow-AF ({\color{blue}ECCV'20}) \cite{yuan2020dlow} proposes a sampling method that maps with a set of co-related latent codes, in order to improve the sample diversity for a CVAE-based generative model.
    \item LDS-AF ({\color{blue}ICCV'21}) \cite{ma2021likelihood} is a flow-based model that uses likelihood-based diverse sampling to improve the diversity of the multi-path prediction.
    \item AgentFormer ({\color{blue}ICCV'21}) \cite{yuan2021agentformer} employs attention mechanisms to model interactions among agents and a CVAE-based model to generate multi-paths for each agents.
    \item CoverNet ({\color{blue}CVPR'20}) \cite{phan2020covernet} is the baseline model of nuScenes. It formulates the multi-path prediction problem as a classification over the states of a trajectory and combines a CNN-based scene extractor to generate diverse predictions.
    \item Trajetron++ ({\color{blue}ECCV'20}) \cite{salzmann2020trajectron++} incorporates agent's motion information and dynamic scene information to train a graph-structured CVAE-based model for trajectory forecasting.
\end{itemize}

Besides AgentFormer and Trajetron++, we benchmark a set of various models on the ETH/UCY datasets for pedestrian trajectory prediction.
\begin{itemize}
    \item Social GAN ({\color{blue}CVPR'18}) \cite{gupta2018social} is a GAN-based trajectory predictor. It aggregates the hidden states of the agents to learn their interactions.
    \item SoPhie ({\color{blue}CVPR'18}) \cite{sadeghian2019sophie} incorporates attentions to extract scene information for a GAN-based model for trajectory prediction.
    \item STAR ({\color{blue}ECCV'20}) \cite{yu2020spatio} is a spatio-temporal graph transformer-based model for predicting pedestrian trajectories. 
    \item MID ({\color{blue}CVPR'22})\cite{gu2022stochastic} is the first diffusion model that effectively models pedestrian stochastic trajectories.
    \item LB-EBM ({\color{blue}CVPR'22}) \cite{pang2021trajectory} is a latent belief energy-based model that is learned from expert demonstrations.
    \item PCCSNet ({\color{blue}ICCV'21}) \cite{sun2021three} formulates the multimodal trajectory prediction problem into three steps: modality clustering, classification and prediction with modality synthesis.
    \item ScePT ({\color{blue}CVPR'22}) \cite{chen2022scept} is a policy planning-based model that can generate scene-consistent trajectories.
    \item MemoNet ({\color{blue}CVPR'22}) \cite{xu2022remember} is an instant-based model that predicts agents' motion by searching for similar scenarios in the training data.
    \item GP-Graph ({\color{blue}ECCV'22}) \cite{bae2022learning} models both individual-wise and group-wise relations as graph representations and provides collective group representations for pedestrian trajectory prediction in crowded environments.
    \item SHENet ({\color{blue}Neurips'22}) \cite{meng2022forecasting} categorizes scene history information into historical group trajectory and individual-surroundings interaction, which is leveraged to guide the prediction in similar scenarios.
    \item EqMotion ({\color{blue}CVPR'23}) \cite{xu2023eqmotion} proposes an equivariant geometric feature learning module to learn an Euclidean transformable feature for invariant interaction reasoning.
    \item LED ({\color{blue}CVPR'23}) \cite{mao2023leapfrog} is a diffusion-based model that uses a leapfrog initializer to directly learn future trajectory's multimodal distribution.
\end{itemize}

\subsection{Experimental Setting}
\label{sec:setting}
The hyper-parameters of GATraj are as follows.
The hidden states and the embedding dimensions are all set to 64.
We apply the Adam optimizer \cite{kingma2015adam} with a learning rate of $5e^{-4}$ and a cosine annealing schedule until it reaches $1e^{-5}$.
The batch size is set to 32, and the maximum epoch is set to 1000.
On ETH/UCY, the maximum distance of the ego and neighboring agents is set to \SI{10}{m} and the GCN layers for messaging passing are set to $l=2$.
On nuScenes, given the faster driving speed of vehicles and fewer neighboring agents than ETH/UCY, the maximum distance of the ego and neighboring agents is set to \SI{100}{m} and the GCN layers for messaging passing are set to $l=1$. 
All our models are trained on \href{https://colab.research.google.com/}{Google Colab} with a single Tesla P100 GPU. 
Our code is released at \url{https://github.com/mengmengliu1998/GATraj} with more detailed settings.

\subsection{Results}
\label{results}

\subsubsection{Quantitative Results}
\label{QuantitativeResults}
First, we present the quantitative prediction results on the nuScenes benchmark. 
As the map context information is a strong prior to constrain vehicle trajectories driving on the lane of road \cite{nuscenes}, we compare our model to the other models without any map information and with map information, separately.
As shown in Table~\ref{tab:nuscenesbaseline1}, our model GAtraj achieves the best performance measured by ADE and FDE in predicting five and ten trajectories for each agent, compared to the CVAE-based models DLow-AF and AgentFormer w/o map information, as well as the flow-based model LDS-AF.
GAtraj reduces the prediction errors around ten centimeters compared to the runner up model AgentFormer w/o map information.
Interestingly, as shown in Table~\ref{tab:nuscenesbaseline2}, when comparing GAtraj with the map-based models, it still maintains a relatively good level of prediction performance. 
For example, it surpasses the convolutional-based baseline CoverNet with a clear margin and performs slightly better than the CAVE-based Trajetron++ with scene context information to constrain the predictions.
Moreover, GAtraj only marginally falls behind AgentFormer measured in ADE, even though the latter benefits largely by including the scene information.  

\begin{table}[hbpt!]
\centering
\small
\caption{The quantitative results on nuScenes \cite{nuscenes} compared to the models without using map information. The best/2nd best performances are indicated in boldface and underline.}
\label{tab:nuscenesbaseline1}
\setlength{\tabcolsep}{3pt}
\begin{tabular}{lccccc}
\toprule
Model                 & Map info.    & ADE$_5$ & FDE$_5$ & ADE$_{10}$ & FDE$_{10}$ \\ \midrule
DLow-AF\,\cite{yuan2020dlow}   & No          & 2.11    & 4.70    & 1.78        &3.58         \\
LDS-AF\,\cite{ma2021likelihood}  & No           & 2.06    & 4.62    & 1.65        &3.50         \\
AgentFormer\,\cite{yuan2021agentformer} & No         & \underline{1.97}       &  \underline{4.21}      &  \underline{1.58}       & \underline{3.14}         \\   
GATraj (Ours)            & No         & \textbf{1.87}        & \textbf{4.08}       & \textbf{1.46}        & \textbf{2.97}         \\
\bottomrule   
\end{tabular}
\end{table}

\begin{table}[hbpt!]
\centering
\small
\caption{The quantitative results on nuScenes \cite{nuscenes} compared to the models with using map information. The best/2nd best performances are indicated in boldface and underline.}
\label{tab:nuscenesbaseline2}
\setlength{\tabcolsep}{3pt}
\begin{tabular}{lccccc}
\toprule
Model                 & Map info.    & ADE$_5$ & FDE$_5$ & ADE$_{10}$ & FDE$_{10}$ \\ \midrule
MTP\,\cite{cui2019multimodal}  & Yes          & 2.93    &-        & 2.93        &-         \\
MultiPath\,\cite{chai2019multipath} & Yes     & 2.32    &-        & 1.96        &-         \\
CoverNet baseline\,\cite{nuscenes} &  Yes      & 1.96    &-        & 1.48        &-         \\
Trajetron++ \,\cite{salzmann2020trajectron++} & Yes     & 1.88        &-        & 1.51        & -        \\
AgentFormer\,\cite{yuan2021agentformer} & Yes  & \textbf{1.86}  & \textbf{3.89}  & \textbf{1.45}  & \textbf{2.86}         \\   
GATraj (Ours)            & No         & \underline{1.87}        & \underline{4.08}       & \underline{1.46}        & \underline{2.97}         \\
\bottomrule   
\end{tabular}
\end{table}

Next, we present the quantitative prediction results on ETH/UCY.
Different from the nuScenes dataset with most of the agents being vehicles, 
the agents in the ETH/UCY datasets are all pedestrians. 
As shown in Figure \ref{fig:example2}, pedestrians move freely because the scenes depicted in ETH/UCY consist of open spaces with few constraints on the movements of pedestrians.
Therefore, in these datasets, only a few models make use of scene information, such as SoPhie and ScePT.
To simplify the comparison, we compare GATraj with the other models in Table \ref{tab:ethbaseline} without differentiating the inclusion of scene information \footnote{The results of Trajectron++ and AgentFormer are updated according to implementation issue \#53 \cite{salzmann2020trajectron++github} and issue \#5 \cite{yuan2021agentformergithub}, respectively.}.

\begin{table}[hbpt!]
\centering
\small
\caption{Quantitative results on ETH/UCY \cite{pellegrini2009you,lerner2007crowds} measured by $\text{ADE}_\text{20}$/$\text{FDE}_\text{20}$. The best/2nd best performances are indicated in boldface and underline.}
\label{tab:ethbaseline}
\setlength{\tabcolsep}{2pt}
\begin{tabular}{l|c|c|c|c|c|c}
\toprule
Models           & Eth       & Hotel     & Uni       & Zara1     & Zara2     & Avg   \\ \midrule 
Social GAN\cite{gupta2018social}  & 0.81/1.52 & 0.72/1.61 & 0.60/1.26 & 0.34/0.69 & 0.42/0.84 & 0.58/1.18 \\
{SoPhie}\,\cite{sadeghian2019sophie}   & 0.70/1.43 & 0.76/1.67 & 0.54/1.24 & 0.30/0.63 & 0.38/0.78 & 0.54/1.15 \\
\footnotesize{Trajectron++}\cite{salzmann2020trajectron++}  & 0.67/1.18 & 0.18/0.28 & 0.30/0.54 & 0.25/0.41 & 0.18/0.32 & 0.32/0.55 \\
STAR\,\cite{yu2020spatio}     & 0.36/0.65 & 0.17/0.36 & 0.31/0.62 & 0.26/0.55 & 0.22/0.46 & 0.26/0.53 \\
\footnotesize{AgentFormer}\cite{yuan2021agentformer}  & 0.45/0.75 & 0.14/0.22 & 0.25/0.45 & 0.18/0.30 & 0.14/0.24 & 0.23/0.39 \\
MID\,\cite{gu2022stochastic}     & 0.39/0.66 & 0.13/0.22 & 0.22/0.45 & 0.17/0.30 & \underline{0.13}/0.27 & 0.21/0.38 \\
LB-EBM\,\cite{pang2021trajectory}    & 0.30/\underline{0.52} & 0.13/0.20 & 0.27/0.52 & 0.20/0.37 & 0.15/0.29 & 0.21/0.38 \\
PCCSNet\,\cite{sun2021three}    & 0.28/0.54 & \underline{0.11}/0.19 & 0.29/0.60 & 0.21/0.44 & 0.15/0.34 & 0.21/0.42 \\
{ScePT}\,\cite{chen2022scept}   & \textbf{0.10}/0.65 & 0.13/0.77 & \textbf{0.12}/0.65 & \textbf{0.13}/0.77 & 0.14/0.81 & \textbf{0.12}/0.73 \\
{MemoNet}\,\cite{xu2022remember}  & 0.40/0.61 & \underline{0.11}/\underline{0.17} & 0.24/0.43 & 0.18/0.32 & 0.14/0.24 & 0.21/0.35 \\
{GP-Graph}\,\cite{bae2022learning} & 0.43/0.63 & 0.18/0.30 & 0.24/\underline{0.42} & 0.17/0.31 & 0.15/0.29 & 0.23/0.39 \\
SHENet\,\cite{meng2022forecasting} & 0.41/0.61 & 0.13/0.20 & 0.25/0.43 & 0.21/0.32 & 0.15/0.26 & 0.23/0.36 \\
{EqMotion}\,\cite{xu2023eqmotion}  & 0.40/0.61 & 0.12/0.18 & 0.23/0.43 & 0.18/0.32 & \underline{0.13}/0.23 & 0.21/0.35 \\
{LED}\,\cite{mao2023leapfrog} & 0.39/0.58 & \underline{0.11}/\underline{0.17} & 0.26/0.43 & 0.18/\textbf{0.26} & \underline{0.13}/\underline{0.22} & 0.21/\underline{0.33} \\
GATraj (Ours)  & \underline{0.26}/\textbf{0.42} & \textbf{0.10}/\textbf{0.15} & \underline{0.21}/\textbf{0.38} & \underline{0.16}/\underline{0.28} & \textbf{0.12}/\textbf{0.21} & \underline{0.17}/\textbf{0.29}  \\\bottomrule
\end{tabular}
\end{table}

Overall, our model achieves the best performance measured in both ADE and FDE on Hotel and Zara2, and the best FDE on Eth, Uni and the average FDE across the datasets for predicting 20 trajectories for each agent.
It achieves the second best on the other datasets, slightly falling behind ScePT in terms of ADE on Eth, Uni, and the average ADE across the datasets.
ScePT uses a conditional policy learning to decode scene-consistent predictions to reduce the average prediction errors, whereas GATraj only conditions on the observed trajectories for prediction.
In addition, due to the auto-regressive policy and the partitioning of the scene-graph into cliques, ScePT requires significantly much longer computational time (see Table~\ref{tab:predictionspeed}).
Compared to the most recent diffusion-model based predictor LED, the performance of GATraj is superior on most datasets, except on Zara1 in terms of FDE. 

\begin{table}[hbpt!]
\centering
\small
\caption{Model efficiency in terms of number of parameters and inference speed on the Zara2 dataset \cite{lerner2007crowds}. The best performance is in boldface.}
\vspace{-11pt}
\label{tab:predictionspeed}
\begin{tabular}{l|cc}
\toprule
Models       & \# Params (K) & Speed (ms) \\ \midrule
STAR \cite{yu2020spatio} &965           &123.2            \\
Trajectron++ \cite{salzmann2020trajectron++} & \textbf{126}           &29.1            \\
AgentFormer \cite{yuan2021agentformer}   & 592          &133.3            \\
PCCSNet \cite{sun2021three} & 347          &81.3  \\
ScePT \cite{chen2022scept}  &    139       & 566.3 \\ \hline
GATraj (Ours) w/o GCN &259    &6.6            \\
GATraj (Ours) w/o GCN, SA &183    & \textbf{3.9}            \\
GATraj (Ours)   &276    &10.1            \\ \bottomrule
\end{tabular}
\vspace{-11pt}
\end{table}
Furthermore, Figure~\ref{fig:inference_time} and Table~\ref{tab:predictionspeed} show the comparison of the model size and the prediction speed between our model and the models with similar prediction performance. Because the settings of the model across the datasets are similar, we conduct this experiment on Zara2 \cite{lerner2007crowds} using a single Tesla V100 GPU and a batch size of 32 to demonstrate its inference speed.
Our full model demonstrates an almost three-times faster prediction speed (\ie~\SI{10.1}{ms} vs. \SI{29.1}{ms}) than the 2nd fastest model Trajectron++.
By removing the GCN-based interaction module and the attention mechanisms, our model can reach a speed of \SI{3.9}{ms}, and the size of our model is also decreased from \SI{276}{K} to \SI{183}{K}. 
Later, we will demonstrate that even without the GCN and attention mechanisms, our model maintains a relatively good prediction accuracy, as shown in Tables~\ref{ablation-nuscenesSAGCN} and \ref{tab:ablation-ethUCYSAGCN}.

\subsubsection{Qualitative Results}
\label{QualitativeResults}
In the following, we present the visualization of the prediction results by GATraj on nuScenes for autonomous driving and ETH/UCY for pedestrian trajectories.

\begin{figure}[t!]
    \centering
    \begin{subfigure}{0.6\linewidth}
    \includegraphics[clip=true, trim=17.65cm 1cm 0cm 0cm, width=1\textwidth]{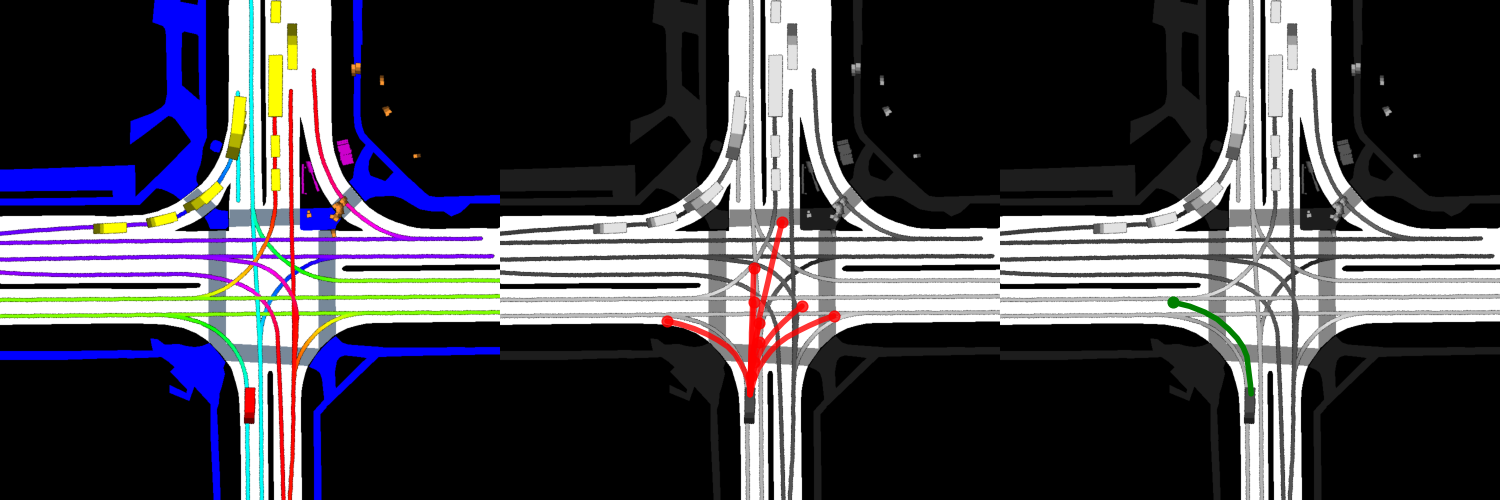}
    \caption{} 
    \end{subfigure}%

    \begin{subfigure}{0.6\linewidth}
    \includegraphics[clip=true, trim=17.65cm 1cm 0cm 0cm, width=1\textwidth]{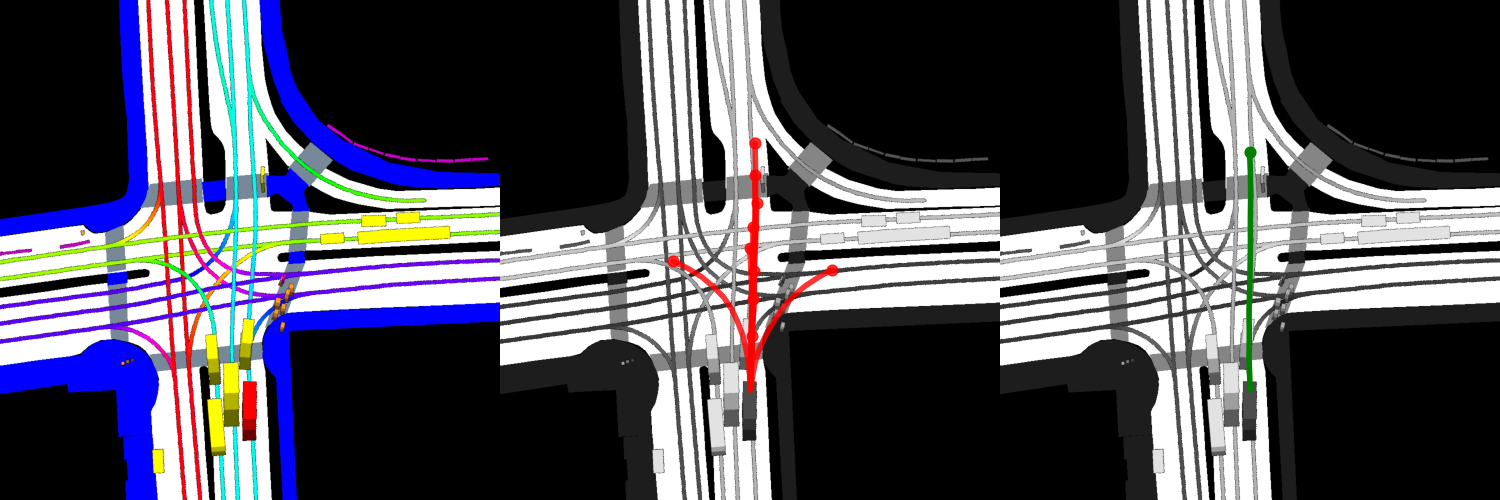}
    \caption{}
    \end{subfigure}%

    \begin{subfigure}{0.6\linewidth}
    \includegraphics[clip=true, trim=17.65cm 1cm 0cm 0cm, width=1\textwidth]{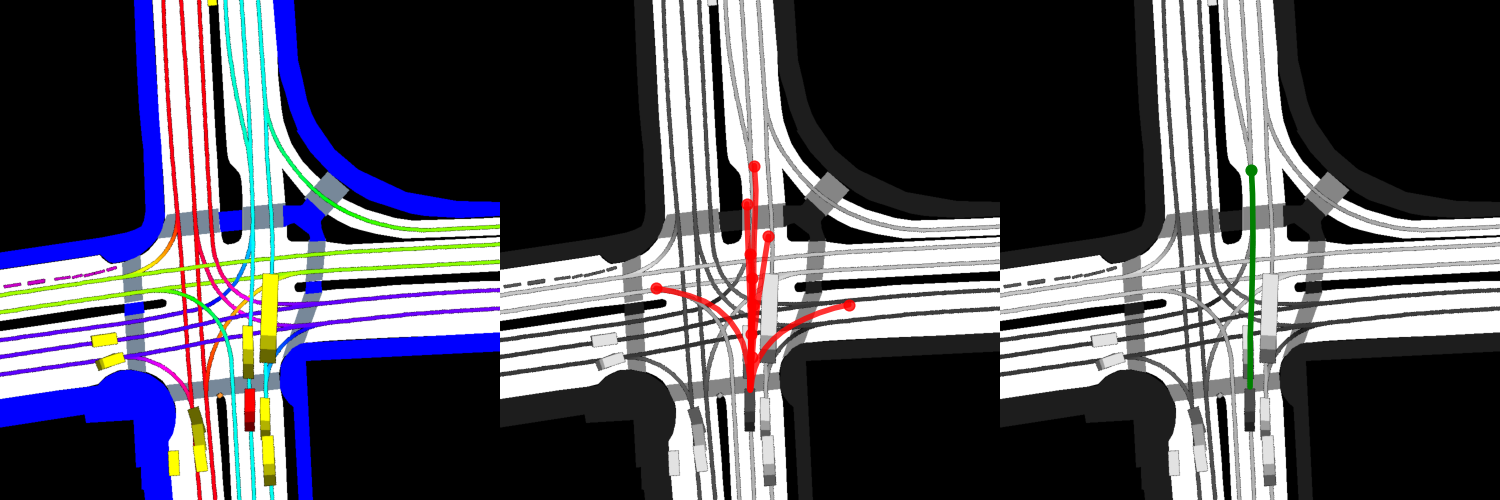}
    \caption{}
    \end{subfigure}%
    
    \begin{subfigure}{0.6\linewidth}
    \includegraphics[clip=true, trim=17.65cm 1cm 0cm 0cm, width=1\textwidth]{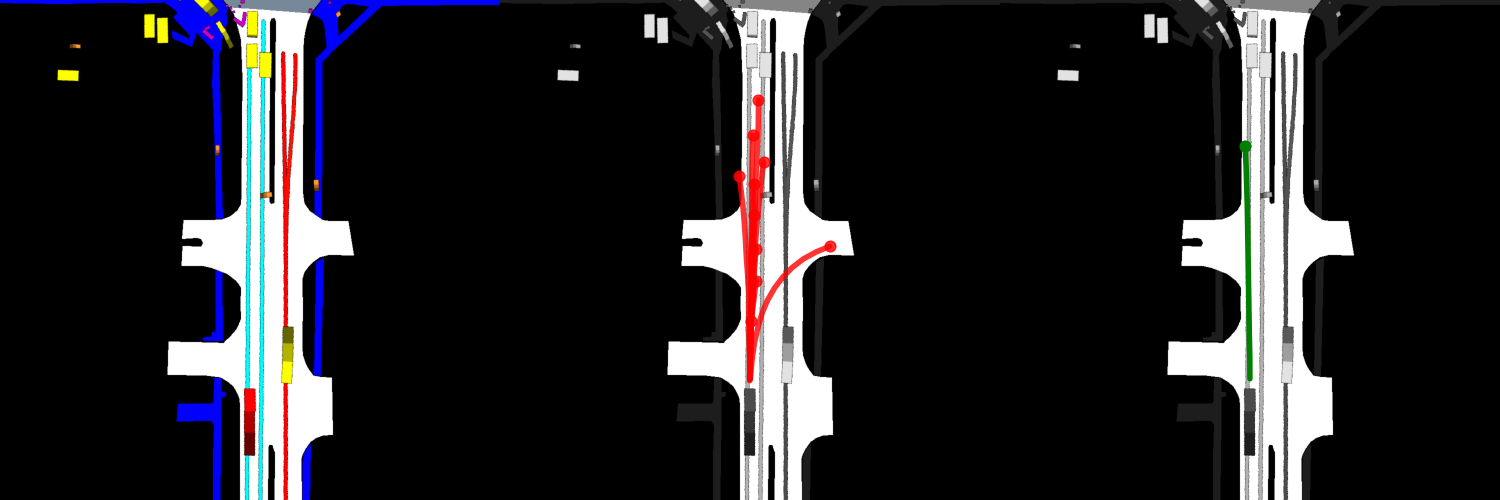}
    \caption{}
    \end{subfigure}%
    \vspace{-11pt}
    \caption{Qualitative results of GATraj on the nuScenes \cite{nuscenes} dataset. Left: predictions, right: the ground truth. The observed trajectory is denoted as dark rectangles with descending grayscale along the time steps -- a darker color indicates an earlier time step. The prediction is in red dotted lines, and the corresponding ground truth is in green dotted lines. The HD map is only used for visualization and not used as extra contextual information for prediction.}
    \label{fig:example1}
    \vspace{-11pt}
\end{figure}

To demonstrate the capability of predicting multimodal trajectories of the target vehicle in nuScenes, we visualize ten potential trajectories ($K=10$) in each scenario, as shown in Figure~\ref{fig:example1}. GATraj generates diverse multimodal predictions for the target vehicle, such as: turning into different directions while approaching the intersection in (a) and (b), driving through the intersection in (c), and moving at different speeds while passing a straight road in (d).

\begin{figure}[t!]
    \centering
    \begin{subfigure}{0.32\linewidth}
    \includegraphics[clip=true, trim=0cm 0cm 0cm 0cm, width=0.98\textwidth]{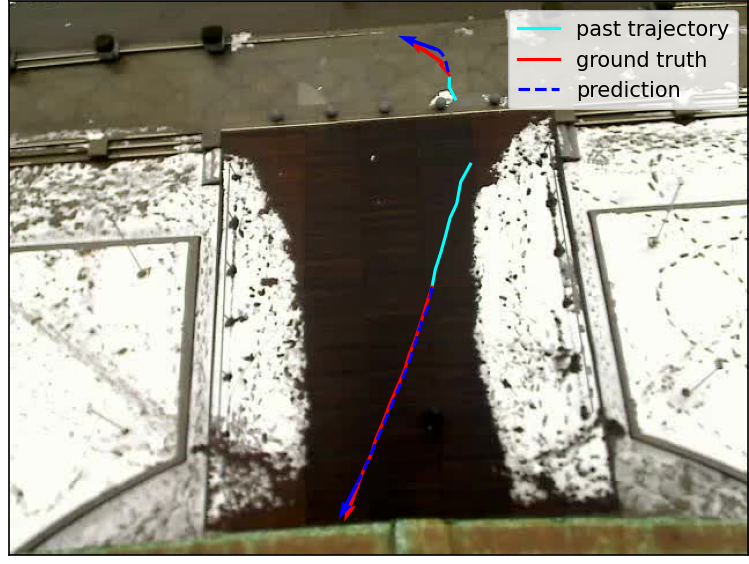}
    \caption{}
    \end{subfigure}%
    \begin{subfigure}{0.32\linewidth}
    \includegraphics[clip=true, trim=0cm 0cm 0cm 0cm, width=0.98\textwidth]{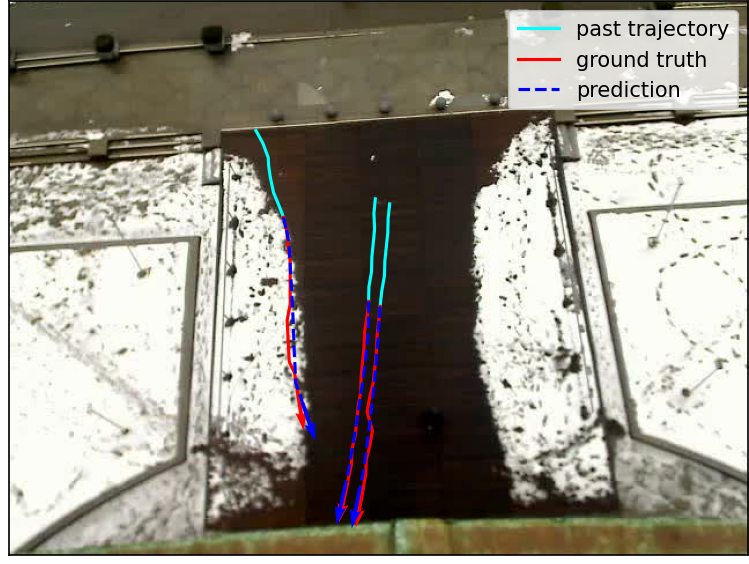}
    \caption{}
    \end{subfigure}%
    \begin{subfigure}{0.32\linewidth}
    \includegraphics[clip=true, trim=0cm 0cm 0cm 0cm, width=0.98\textwidth]{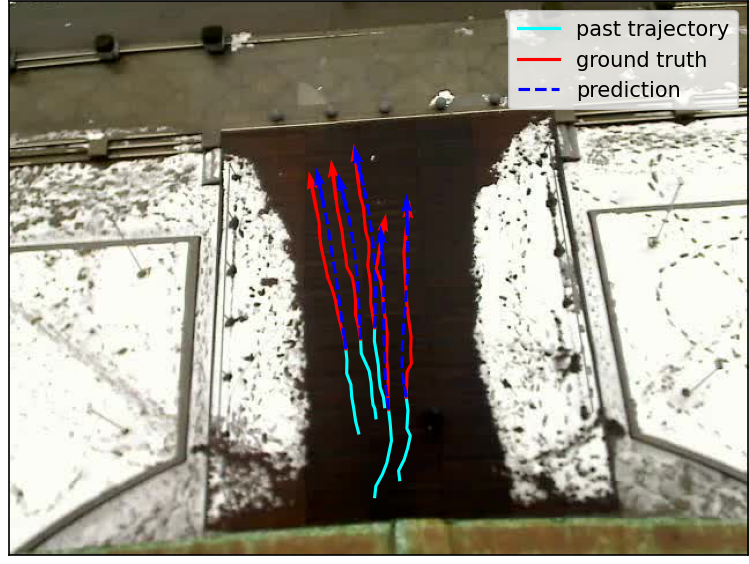}
    \caption{}
    \end{subfigure}%

    \begin{subfigure}{0.32\linewidth}
    \includegraphics[clip=true, trim=0cm 0cm 0cm 0cm, width=0.98\textwidth]{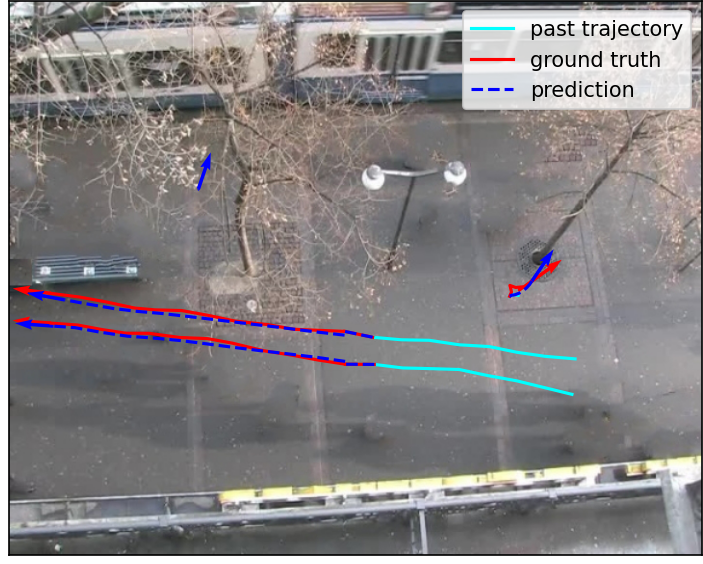}
    \caption{}
    \end{subfigure}%
    \begin{subfigure}{0.32\linewidth}
    \includegraphics[clip=true, trim=0cm 0cm 0cm 0cm, width=0.98\textwidth]{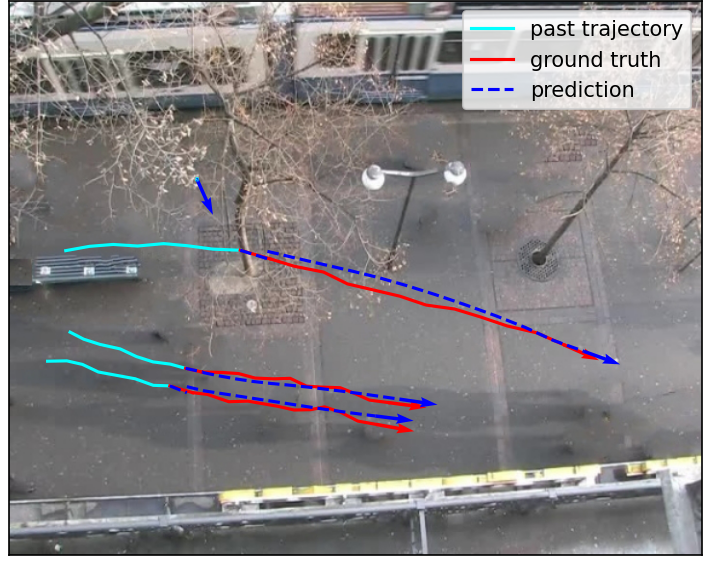}
    \caption{}
    \end{subfigure}%
    \begin{subfigure}{0.32\linewidth}
    \includegraphics[clip=true, trim=0cm 0cm 0cm 0cm, width=0.98\textwidth]{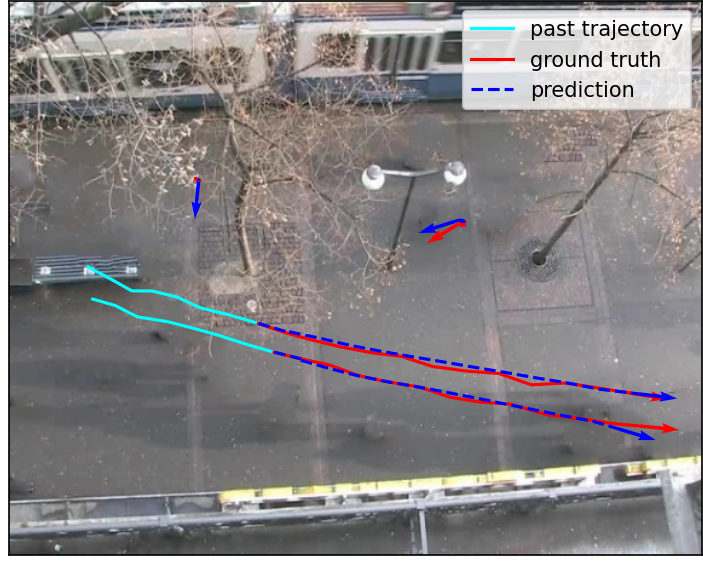}
    \caption{}
    \end{subfigure}%

    \begin{subfigure}{0.32\linewidth}
    \includegraphics[clip=true, trim=0cm 0cm 0cm 0cm, width=0.98\textwidth]{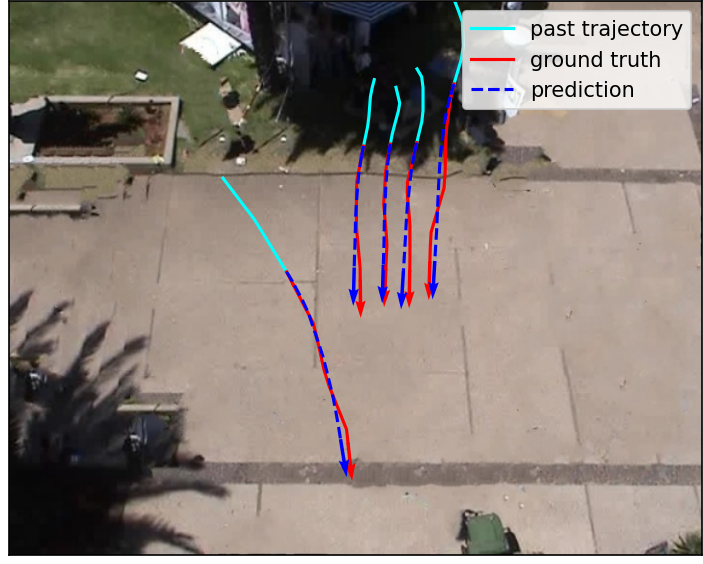}
    \caption{}
    \end{subfigure}%
    \begin{subfigure}{0.32\linewidth}
    \includegraphics[clip=true, trim=0cm 0cm 0cm 0cm, width=0.98\textwidth]{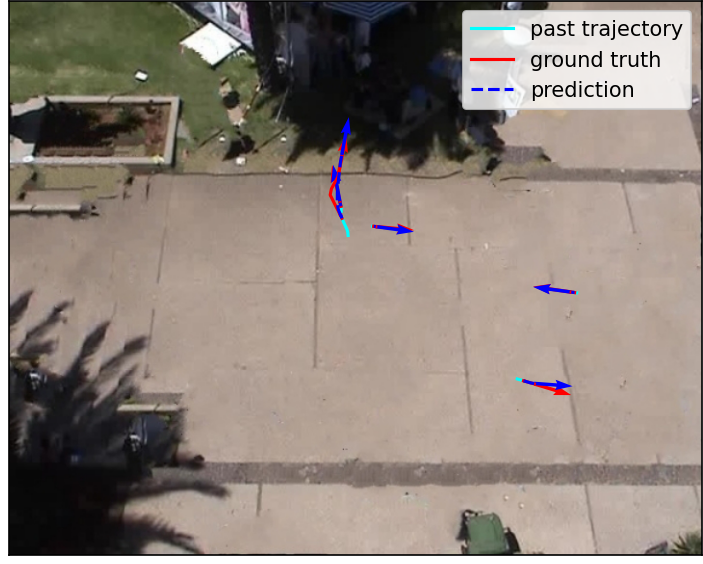}
    \caption{}
    \end{subfigure}%
    \begin{subfigure}{0.32\linewidth}
    \includegraphics[clip=true, trim=0cm 0cm 0cm 0cm, width=0.98\textwidth]{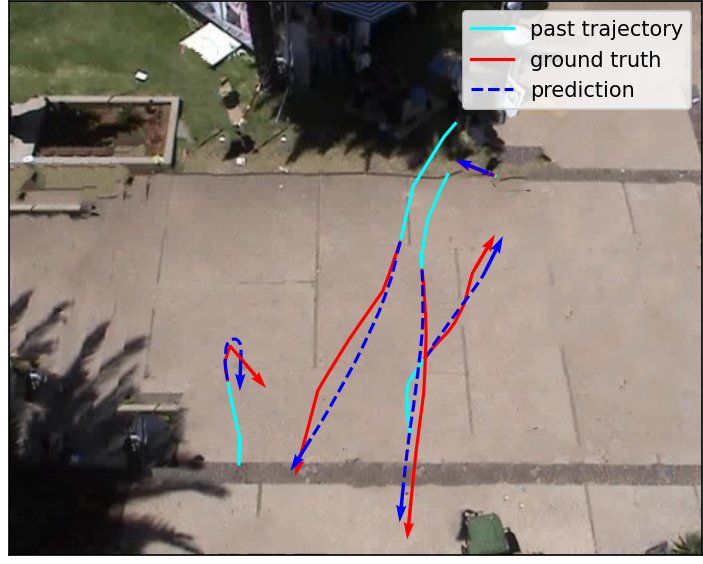}
    \caption{}
    \end{subfigure}%

    \begin{subfigure}{0.32\linewidth}
    \includegraphics[clip=true, trim=0cm 0cm 0cm 0cm, width=0.98\textwidth]{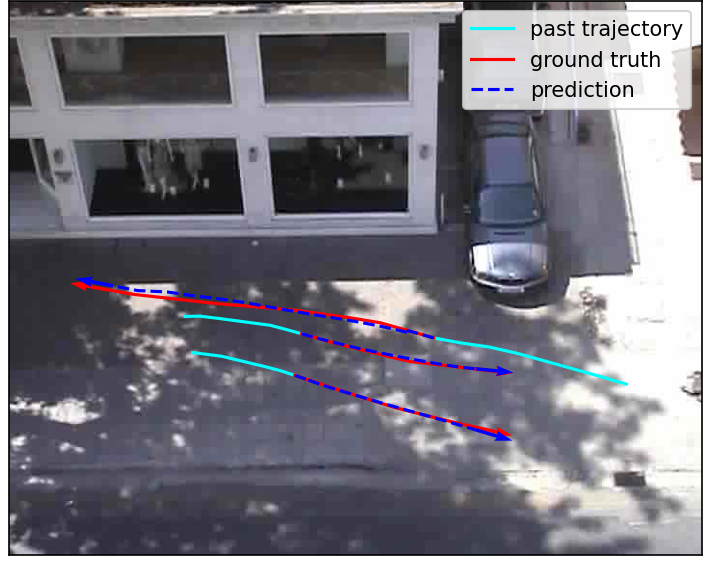}
    \caption{}
    \end{subfigure}%
    \begin{subfigure}{0.32\linewidth}
    \includegraphics[clip=true, trim=0cm 0cm 0cm 0cm, width=0.98\textwidth]{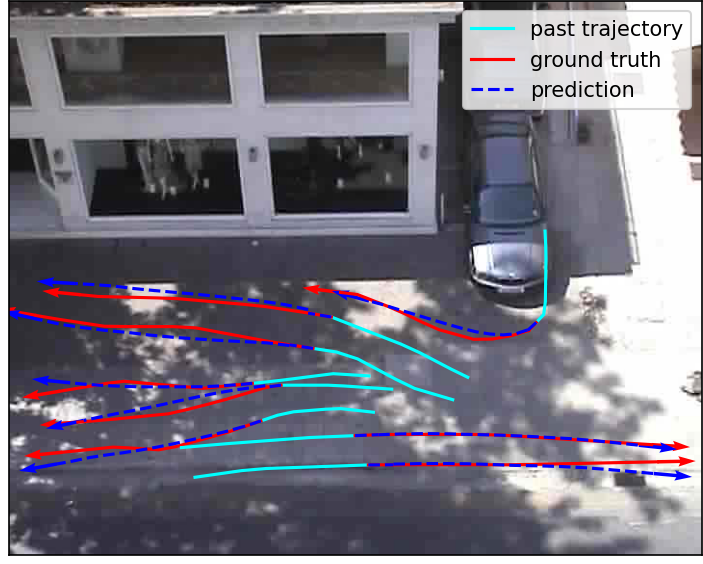}
    \caption{}
    \end{subfigure}%
    \begin{subfigure}{0.32\linewidth}
    \includegraphics[clip=true, trim=0cm 0cm 0cm 0cm, width=0.98\textwidth]{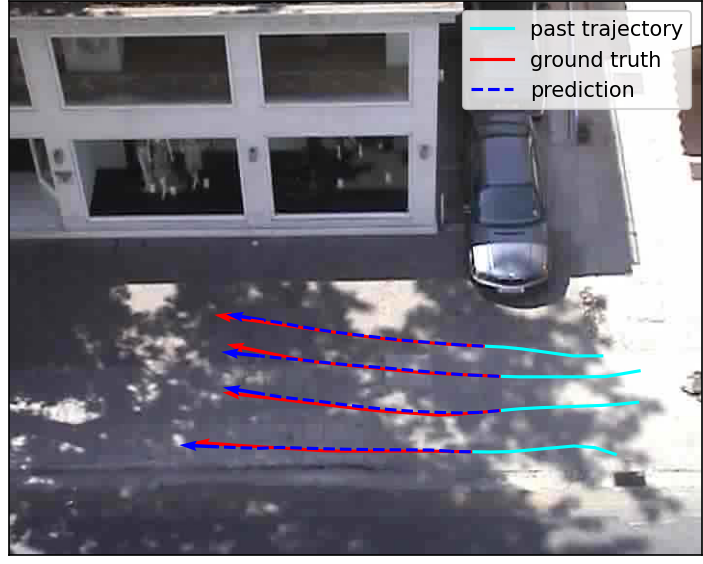}
    \caption{}
    \end{subfigure}%
    \caption{Qualitative results tested on the ETH/UCY \cite{pellegrini2009you,lerner2007crowds} datasets. From the first row to the forth rows, they are the scenarios in Eth, Hotel, Uni, and Zara1/2.} 
    \vspace{-10pt}
    \label{fig:example2}
\end{figure}

As pedestrians move closely with each other in ETH/NYC, in Figure \ref{fig:example2}, we first only plot the best prediction of 20 trajectories (K=20) for each pedestrian for clear visualization.
The first row shows the scenarios in ETH. GATraj correctly predicts a pedestrian walking diagonally in the passage way and another pedestrian turning around to the street in (a). It also correctly predicts pedestrians walking in parallel in a small group (two pedestrians) in (b) and a big group (five pedestrians) in (c).
Similarly, even though pedestrians move at various speeds, we can see that the predictions of GATraj are well aligned with the ground truth trajectories, especially for the final positions, in the second row for the scenarios in Hotel and the third row for the scenarios in Uni.
Interestingly in (i), GATraj can generate reasonable non-linear predictions, such as turning around or changing direction suddenly.
The last row shows the predictions of the scenarios in Zara01/02. GATraj demonstrates that it can generate good predictions for multiple pedestrians walking closely to each other, and it also correctly predicts a pedestrian curving around a parked vehicle in (k).

Furthermore, in Figure \ref{fig:example2}, we showcase the multimodal predictions for pedestrian trajectory prediction, \ie~20 trajectories for each pedestrian.
It can be observed from the predictions that GATraj is capable of generating highly diverse predictions in terms of moving direction and speed.
It is intriguing to note that some of the predictions capture potential intentions of the pedestrian. For instance, in (b), certain predictions indicate the tram station along the road, and one prediction even suggests that the pedestrian might turn around.
Similarly, in (d), one prediction points towards the shop window, while other predictions indicate different directions as the pedestrian moves further into the intersection of the road.
\begin{figure}[t!]
    \centering
    \begin{subfigure}{0.32\linewidth}
    \includegraphics[clip=true, trim=0cm 0cm 0cm 0cm, width=0.98\textwidth]{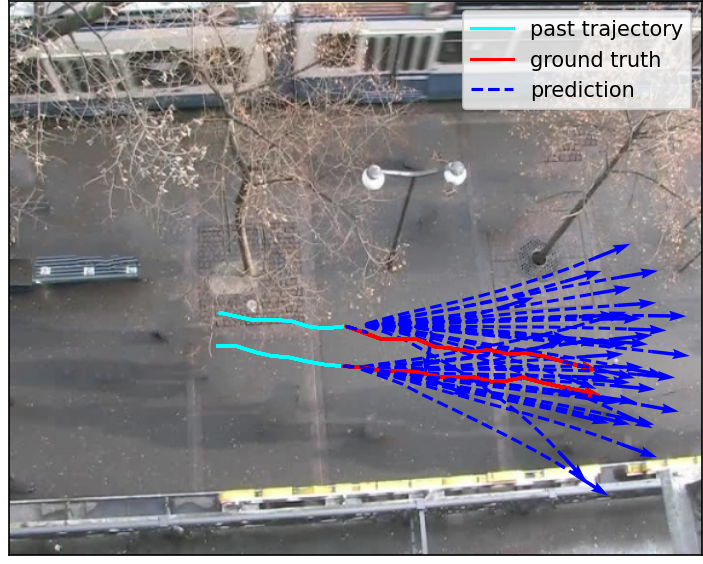}
    \caption{}
    \end{subfigure}%
    \begin{subfigure}{0.32\linewidth}
    \includegraphics[clip=true, trim=0cm 0cm 0cm 0cm, width=0.98\textwidth]{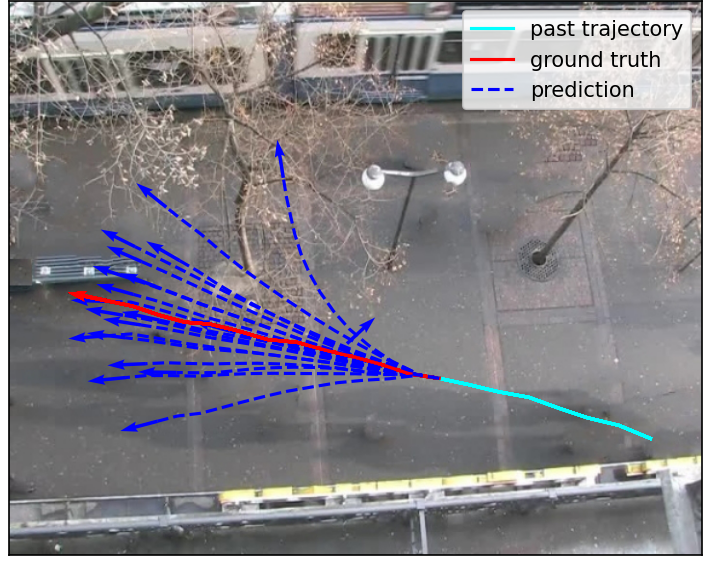}
    \caption{}
    \end{subfigure}%
    \begin{subfigure}{0.32\linewidth}
    \includegraphics[clip=true, trim=0cm 0cm 0cm 0cm, width=0.98\textwidth]{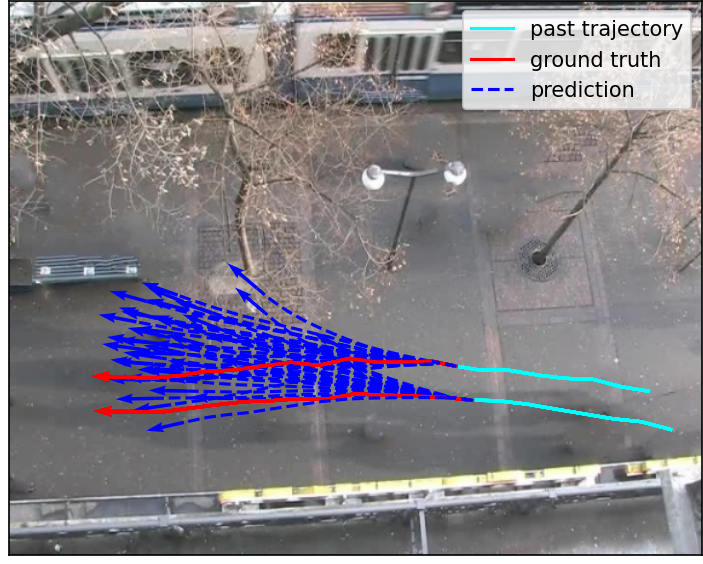}
    \caption{}
    \end{subfigure}%

    \begin{subfigure}{0.32\linewidth}
    \includegraphics[clip=true, trim=0cm 0cm 0cm 0cm, width=0.98\textwidth]{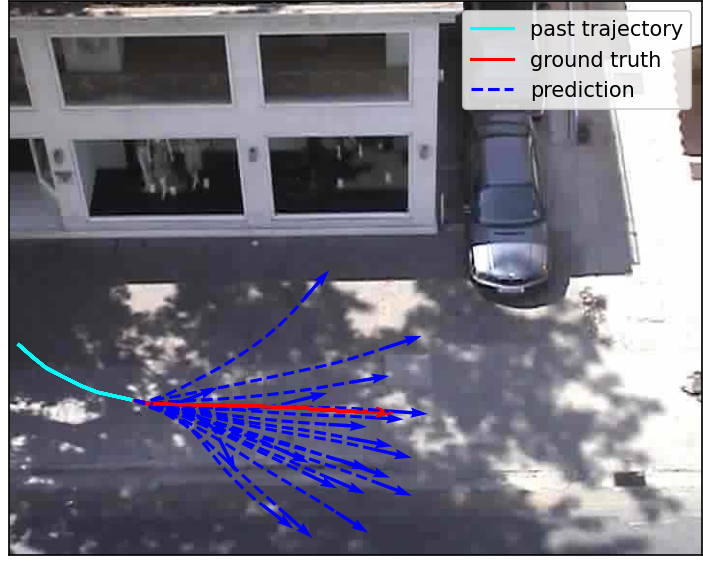}
    \caption{}
    \end{subfigure}%
    \begin{subfigure}{0.32\linewidth}
    \includegraphics[clip=true, trim=0cm 0cm 0cm 0cm, width=0.98\textwidth]{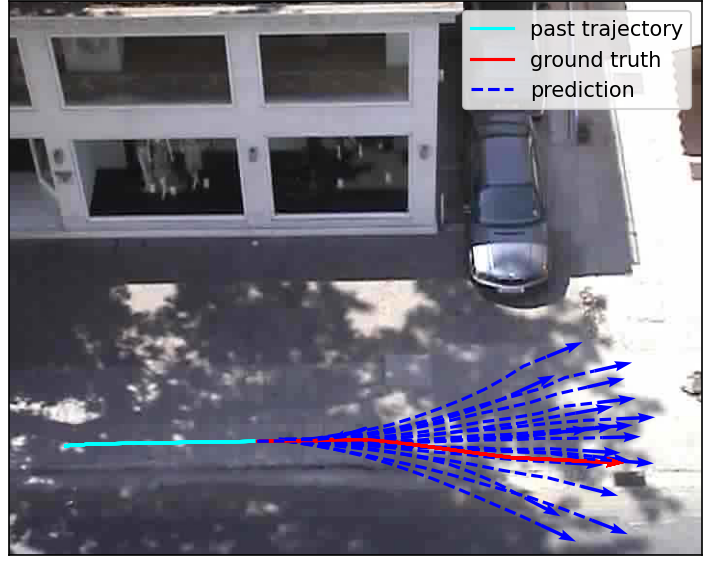}
    \caption{}
    \end{subfigure}%
    \begin{subfigure}{0.32\linewidth}
    \includegraphics[clip=true, trim=0cm 0cm 0cm 0cm, width=0.98\textwidth]{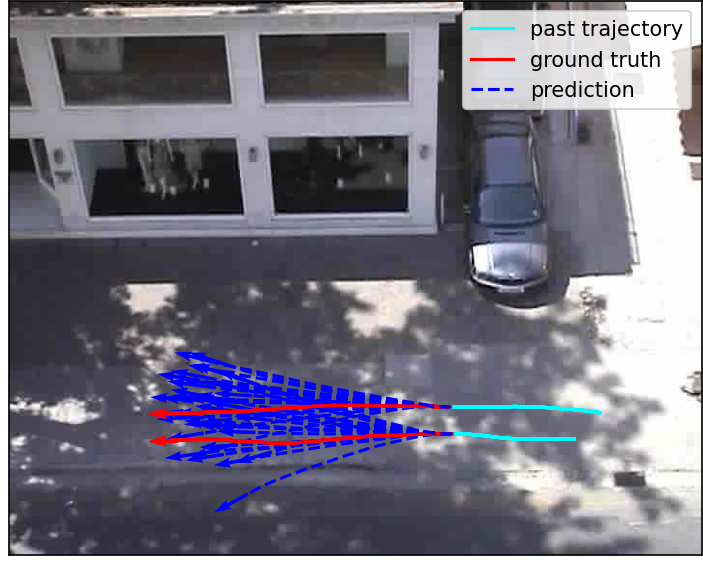}
    \caption{}
    \end{subfigure}%
    \caption{Qualitative results with multimodal predictions tested on the ETH/UCY \cite{pellegrini2009you,lerner2007crowds} datasets.}    
    \label{fig:example3}
\end{figure}

\subsection{Ablation Studies}
\label{ablationstudies}
In this subsection, we carry out a series of ablation studies to provide a comprehensive insight of GATraj's performance.

First, we start with analyzing the performance of the Laplacian MDN decoder (LMM).  
We compare it with a Gaussian MDN decoder (GMM) using negative log-likelihood for the multimodal predictions. 
Both decoders use the same layers of neural network, except for the parameterization of the outputted distributions. 

\begin{figure}[hbpt!]
\begin{center}
 \includegraphics[clip=true, trim=0pt 0pt 0pt 0pt, height=4.5cm]{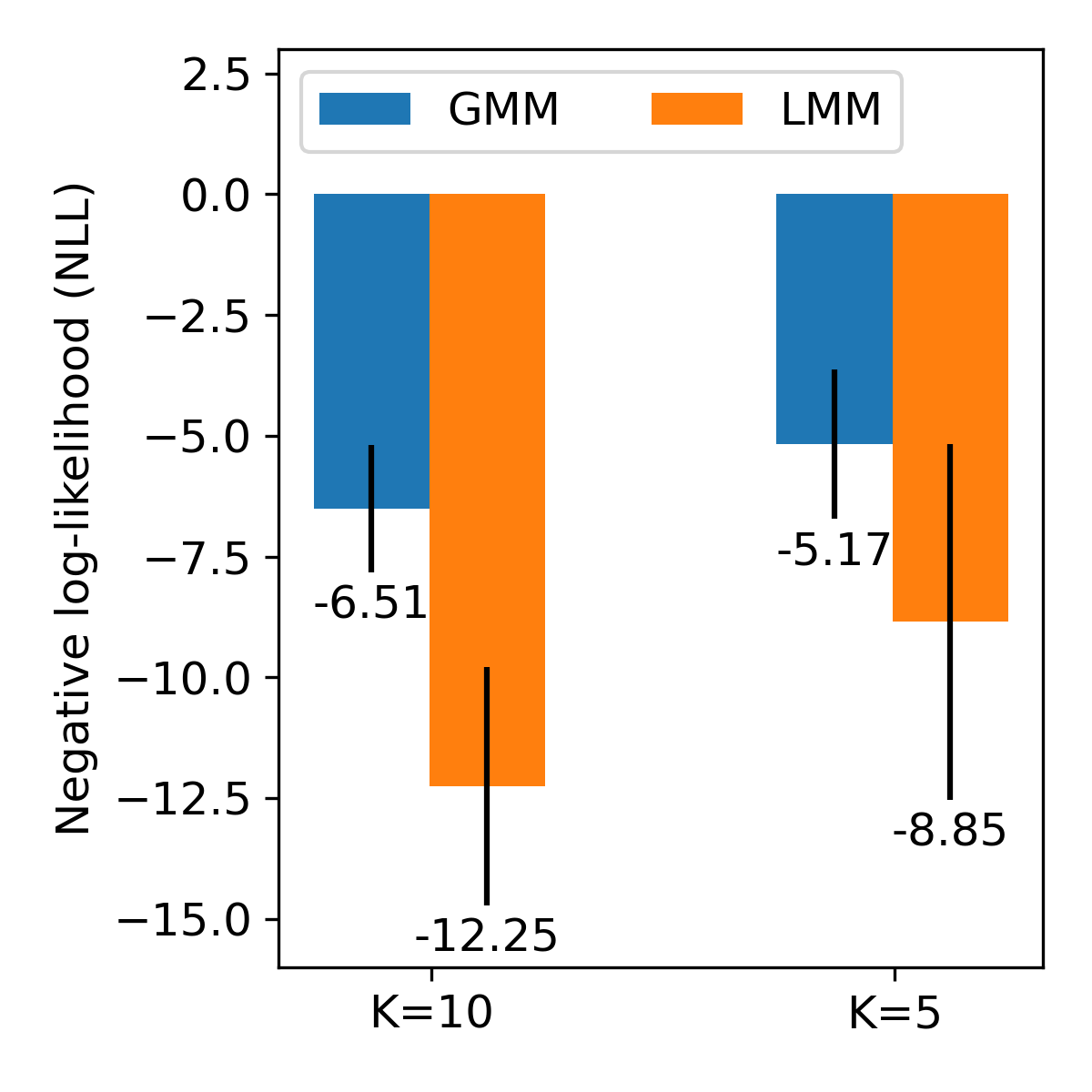}
\end{center}
   \caption{The negative log-likelihood (NLL) between LMM and GMM tested on nuScenes. The error bars stand for the standard deviations.}
\label{fig:nllnuscenes}
\end{figure}

As shown in Figure \ref{fig:nllnuscenes}, it is clear that the LMM decoder achieves a much lower negative log-likelihood (NLL) than the GMM decoder for predicting both ten modes ($K=10$) and five modes ($K=5$) on nuScenes for vehicle trajectory prediction. 
the LMM decoder also achieves evidently lower NLL across all the datasets in ETH/UCY for pedestrian trajectory prediction with 20 modes.

\begin{figure}[hbpt!]
\begin{center}
 \includegraphics[clip=true, trim=0pt 0pt 0pt 0pt, height=4.5cm]{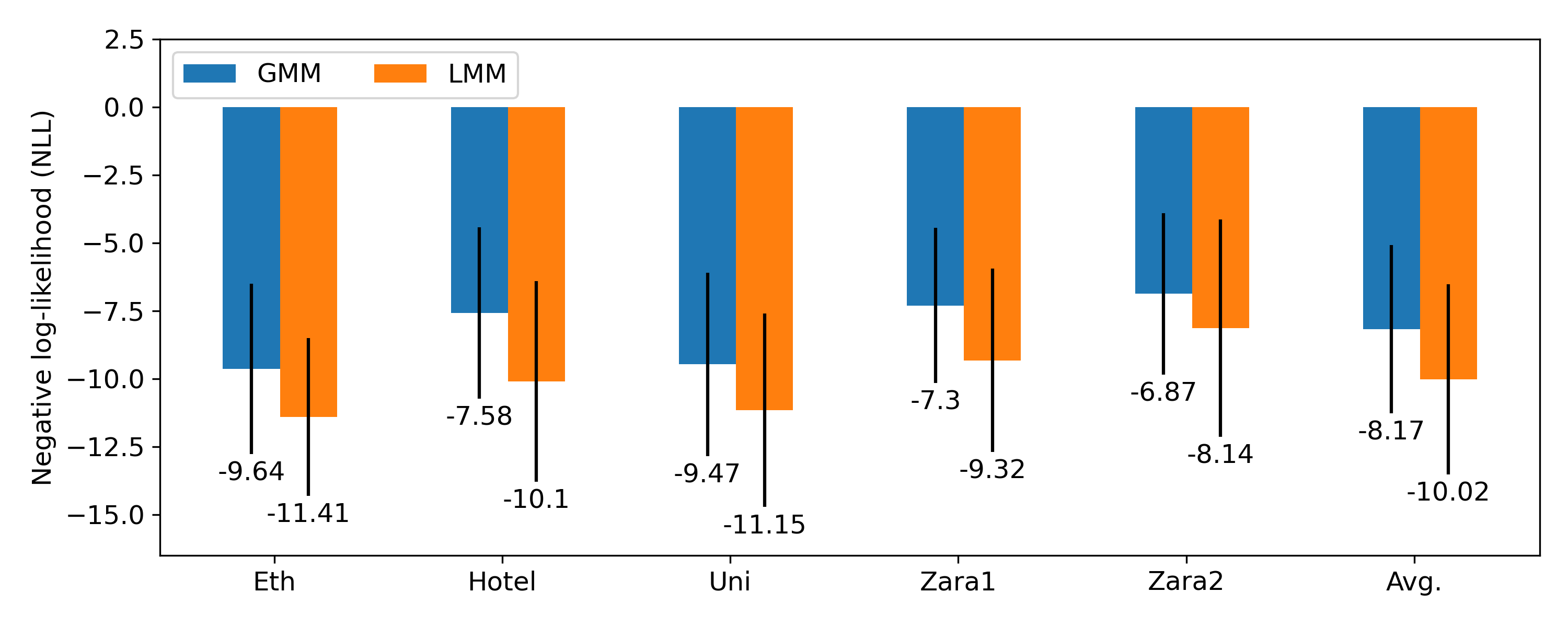}
\end{center}
   \caption{The negative log-likelihood (NLL) between LMM and GMM tested on ETH/UCY. The error bars stand for the standard deviations}
\label{fig:nllethucy}
\end{figure}

Moreover, Tables \ref{ablation-decoder-nuscenes} and \ref{ablation-decoder-ethucy} show the comparison of the predictions errors measured by ADE and FDE on nuScenes and ETH/UCY, respectively, between GMM and LMM using the same network configuration except for the outputted distributions.
Both tables clearly show that the LMM decoder generate better predictions measured in lower ADE and FDE on each dataset.

To summarize, these consistently better results in terms of NLL and prediction errors of LMM compared to GMM validate that the LMM decoder is more accurate in estimating the probability of multimodal predictions.

\begin{table}[hbpt!]
\centering
\small
\caption{The comparison of the prediction errors between LMM and GMM on nuScenes.} 
\setlength{\tabcolsep}{2pt}
\begin{tabular}{ c | c c c c  } 
\toprule
Decoder & $ADE_{5}$  & $FDE_{5}$  & $ADE_{10}$  & $FDE_{10}$  \\ \midrule 
GMM    &2.08& 4.67& 1.64& 3.45\\
LMM    &2.00& 4.46& 1.57& 3.28\\
\bottomrule
\end{tabular}
\label{ablation-decoder-nuscenes}
\end{table}

\begin{table}[hbpt!]
\centering
\small
\caption{The comparison of the prediction errors (ADE/FDE) between LMM and GMM on ETH/UCY.} 
\label{ablation-decoder-ethucy}
\setlength{\tabcolsep}{2pt}
\begin{tabular}{l|c|c|c|c|c|c}
\toprule
Models              & Eth       & Hotel     & Uni       & Zara1     & Zara2     & Avg.      \\ \midrule 
GMM            & 0.27/0.47 & 0.11/0.18 & 0.22/0.40 & 0.18/0.34 & 0.12/0.23 & 0.18/0.32        \\
LMM         & 0.26/0.42 & 0.10/0.15 & 0.21/0.38 & 0.16/0.28 & 0.12/0.21 & 0.17/0.29        \\ \bottomrule
\end{tabular}  
\end{table}

Next, we conduct the ablation study to analyze the effectiveness of the self-attention mechanism (SA) to enhance the learning of spatial-temporal information and the GCN module to model interactions among agents.
The effectiveness of the self-attention mechanisms is not so obvious on nuScenes (as shown in Table \ref{ablation-nuscenesSAGCN}), while the model's performance decreases without the self-attention mechanisms on ETH/UCY (as shown in Table \ref{tab:ablation-ethUCYSAGCN}).
The reason could be that nuScenes only lets the model observe a short trajectory, \ie~from two to four-time steps, while ETH/UCY provides eight-time steps for observation;
The self-attention mechanisms may work better on longer sequences.
After removing the GCN module, the performance drops clearly on both datasets.
\begin{table}[hbpt!]
\centering
\small
\caption{The ablation study of the self-attention (SA) mechanism and the GCN module on nuScenes.} 
\setlength{\tabcolsep}{2.2pt}
\begin{tabular}{c c| c c c c  } 
\toprule
SA      & GCN   & $ADE_{5}$  & $FDE_{5}$  & $ADE_{10}$  & $FDE_{10}$  \\ \midrule 
-       & -     &2.00& 4.45& 1.57& 3.29\\
$\surd$ & -     &2.00& 4.46& 1.57&3.28 \\
$\surd$ &$\surd$& 1.87& 4.08& 1.46&2.97\\\bottomrule
\end{tabular}
\label{ablation-nuscenesSAGCN}
\end{table}

\begin{table}[hbpt!]
\centering
\small
\caption{The ablation study of the self-attention (SA) mechanism and the GCN module on ETH/UCY.}
\label{tab:ablation-ethUCYSAGCN}
\setlength{\tabcolsep}{2pt}
\begin{tabular}{cc|c|c|c|c|c|c}
\toprule
SA      & GCN       & Eth       & Hotel     & Uni       & Zara1     & Zara2     & Avg.      \\ \midrule 
-       & -         & 0.29/0.49 & 0.11/0.17 & 0.22/0.40 & 0.17/0.32 & 0.12/0.22 & 0.18/0.32        \\
$\surd$ & -         & 0.29/0.45 & 0.10/0.15 & 0.21/0.40 & 0.17/0.32 & 0.12/0.22    & 0.18/0.31         \\
$\surd$ &$\surd$    & 0.26/0.42 & 0.10/0.15 & 0.21/0.38 & 0.16/0.28 & 0.12/0.21 & 0.17/0.29  \\ \bottomrule
\end{tabular}  
\end{table}

In addition, we substitute the LSTM layer by an MLP layer in the decoder to analyze the effect of the hidden states in the LSTM for temporal information.
It can be seen from Tables \ref{ablation-nuscenes-lstmmlp} and \ref{tab:ablation-ethUCY-lstmmlp},
replacing the LSTM layer with an MLP in the decoder leads to an apparent decrease in performance on both nuScenes and ETH/UCY.

\begin{table}[hbpt!]
\centering
\small
\caption{The comparison between the LSTM and MLP layers in the decoder on nuScenes.} 
\setlength{\tabcolsep}{2.2pt}
\begin{tabular}{c| c c c c  } 
\toprule
Decoder layer & $ADE_{5}$  & $FDE_{5}$  & $ADE_{10}$  & $FDE_{10}$  \\ \midrule 
LSTM & 2.00& 4.46& 1.57&3.28 \\
MLP  & 2.03& 4.51& 1.58&3.28 \\ 
\bottomrule
\end{tabular}
\label{ablation-nuscenes-lstmmlp}
\end{table}

\begin{table}[hbpt!]
\centering
\small
\caption{The comparison between the LSTM and MLP layers in the decoder on ETH/UCY.}
\label{tab:ablation-ethUCY-lstmmlp}
\setlength{\tabcolsep}{2pt}
\begin{tabular}{c|c|c|c|c|c|c}
\toprule
Decoder layer                 & Eth       & Hotel     & Uni       & Zara1     & Zara2     & Avg.      \\ \midrule 
LSMT & 0.26/0.42 & 0.10/0.15 & 0.21/0.38 & 0.16/0.28 & 0.12/0.21 & 0.17/0.29          \\ 
MLP  & 0.27/0.43 & 0.11/0.16 & 0.21/0.39 & 0.17/0.31 & 0.12/0.22 & 0.18/0.30        \\\bottomrule
\end{tabular}  
\end{table}

Finally, we investigate the effect of different trajectory inputs, namely, position sequences and offset sequences.
As shown in Table \ref{ablation-nuscenes-input} and \ref{tab:ablation-ethUCY-input}, our model using offset sequences achieves better performance than position sequences on ETH/UCY, but similar performance on nuScenes.
Our conjecture is that the offset sequences are less sensitive to the absolute position in a given scene and using the offset sequences can mitigate the domain gaps across different scenes.
It explains why we observe a clear performance drop when substituting the offset sequences by the position sequences in testing on the ETH/UCY datasets using the leave-one-out cross validation.
In addition, combing both position and offset sequences yield similar performances as that of using the offset sequences alone, leading to no joint benefit. 
Therefore, based on the empirical findings, we decide to use the offset sequences as the trajectory input of our model for trajectory prediction. 

\begin{table}[hbpt!]
\centering
\small
\caption{The comparison of different input features on nuScenes.} 
\setlength{\tabcolsep}{2.2pt}
\begin{tabular}{c c| c c c c  } 
\toprule
Position      & Offset   & $ADE_{5}$  & $FDE_{5}$  & $ADE_{10}$  & $FDE_{10}$  \\ \midrule 
$\surd$       & -     & 1.86 & 4.05 & 1.48 & 3.00 \\
- & $\surd$     & 1.87& 4.08& 1.46&2.97 \\
\bottomrule
\end{tabular}
\label{ablation-nuscenes-input}
\end{table}

\begin{table}[hbpt!]
\centering
\small
\caption{The comparison of different input features on ETH/UCY.}
\label{tab:ablation-ethUCY-input}
\setlength{\tabcolsep}{2pt}
\begin{tabular}{cc|c|c|c|c|c|c}
\toprule
Position & Offset   & Eth       & Hotel     & Uni       & Zara1     & Zara2     & Avg.      \\ \midrule 
$\surd$       & -         & 0.29/0.49 & 0.11/0.17 & 0.22/0.40 & 0.17/0.32 & 0.12/0.22 & 0.18/0.32        \\
- & $\surd$    & 0.26/0.42 & 0.10/0.15 & 0.21/0.38 & 0.16/0.28 & 0.12/0.21 & 0.17/0.29 \\
\bottomrule
\end{tabular}  
\end{table}

\begin{figure}[t!]
\centering
\includegraphics[clip=true, trim=6pt 6pt 6pt 6pt, width=0.75\linewidth]{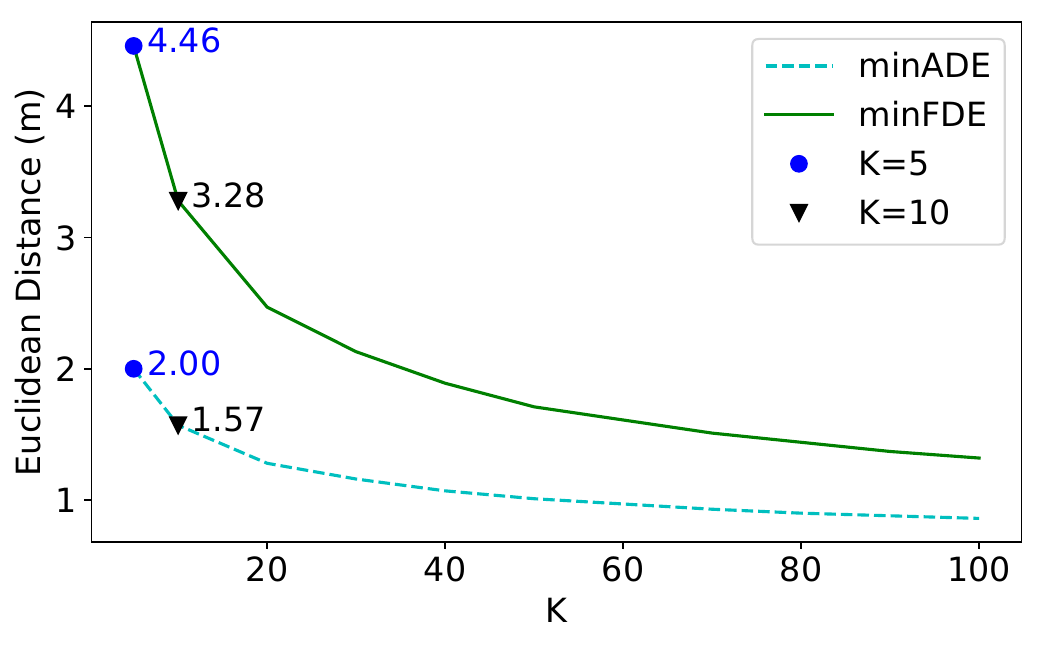}
\caption{The prediction results on nuScenes with increasing the number of components in the Laplacian MDN decoder. GATraj w/o GCN achieves 2.00/4.46 of $\text{ADE}_\text{5}$/$\text{FDE}_\text{5}$ and 1.57/3.28 of $\text{ADE}_\text{10}$/$\text{FDE}_\text{10}$, respectively.}
\label{fig:number_modes}
\end{figure}

We further analyze the performance of GATraj by increasing the number of components. 
The results shown in Fig.~\ref{fig:number_modes} demonstrate that increasing the number of components of the MDN decoder evidently leads to a gain in prediction accuracy.
This trend indicates that the Laplacian MDN decoder can generate diverse multimodal predictions.
However, many benchmarks limit the maximum number of predictions for each agent, \ie~ETH/UCY recommends 20, and nuScene only allows up to ten predictions for each agent.
The limited computational resources of real-time applications may only allow a small number of predictions as well.
This finding and constraint motivate us to work on aggregation strategies to effectively pool out the best prediction in future. 

\subsubsection{Discussion of Limitations and Future Work} 
\label{discussion}
\begin{figure}[hbpt!]
\begin{center}
 \includegraphics[clip=true, trim=17.65cm 0cm 0cm 1cm, width=0.6\linewidth]{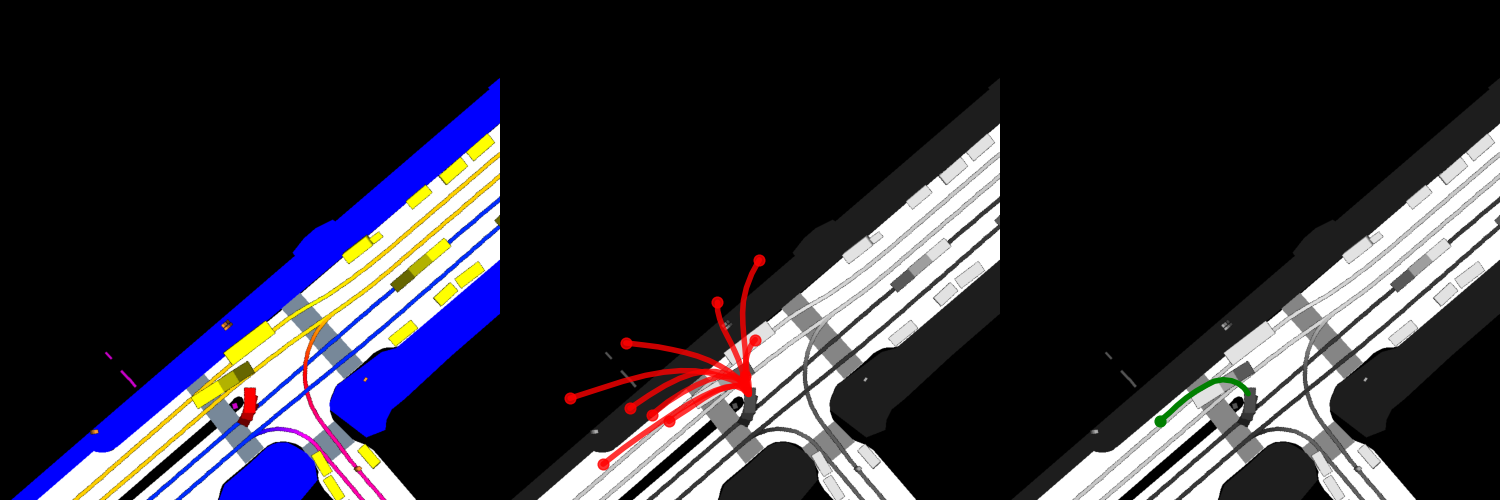}
\end{center}
   \caption{An example of limited performance of GATraj when scene information is not considered. Left: predictions, right: the ground truth. The observed trajectory is denoted as dark rectangles with descending grayscale along the time steps -- a darker color indicates an earlier time step. The prediction is in red dotted lines, and the corresponding ground truth is in green dotted lines. The HD map is only used for visualization and not used as extra contextual information for prediction.}
\label{fig:limitcase}
\end{figure}

Our model does not utilize scene information, making it more challenging to achieve multimodal predictions that are all compliant with the scene. 
This is because the model solely relies on the motion dynamics of agents. 
We assume that all agents behave rationally, such as following scene constraints, and that the observed trajectories include enough data samples that cover various areas within a given scene. 
However, without other explicit scene cues, there is a limitation in associating predictions with unseen areas, which can result in limited performance in achieving consistent predictions that align with the scene.
Figure \ref{fig:limitcase} illustrates this issue, although some predicted trajectories closely match the ground truth trajectory, other predictions deviate from the lanes or violate scene constraints.
GATraj's lack of scene information is the main reason behind its inability to achieve scene-consistent predictions for vehicles driving on dedicated lanes in the nuScenes dataset.

Some recent models in trajectory prediction for autonomous driving, such as DenseTNT \cite{gu2021densetnt}, HiVT \cite{zhou2022hivt}, and QCNet \cite{zhou2023query}, have shown superior performance on trajectory prediction for autonomous driving by exploiting HD maps with lane segment information. 
For these models to function optimally, it is necessary to have access to HD maps that are obtained in advance and are kept up-to-date.
However, the acquisition and maintenance of high-quality and up-to-date HD maps is a costly and time-consuming endeavor.
To address these limitations and enhance generalizability while reducing the burden of HD map annotation and acquisition, perception-based approaches using multiview camera images are gaining more and more attention. 
These approaches generate scene information on-the-fly and project it onto the ground plane from a bird's-eye view, facilitating the trajectory prediction task. 
Most recent examples of such approaches include BEVFormer by Li et al. \cite{li2022bevformer} and planning-based approaches by Hu et al. \cite{hu2023planning}.
In future research, we will explore strategies to incorporate scene information, favorably through perception-based approaches, aiming to achieve scene-compliant multimodal predictions.

\section{Conclusion}
\label{conclusion}
This paper proposes an attention-based graph model named GATraj for multi-agent trajectory prediction with a good balance of prediction accuracy and inference speed.
We use attention mechanisms to learn spatial-temporal dynamics of agents like pedestrians and vehicles and a graph convolutional network to learn scene-centric interactions among them.
A Laplacian mixture density network decoder predicts diverse and multimodal trajectories for each agent.
GATraj achieves performance on par with the state-of-the-art models at a much higher prediction speed tested on the nuScenes benchmark fo autonomous driving and state-of-ther-art performance on the ETH/UCY benchmark for pedestrian trajectory prediction.


\bibliography{main.bib}

\end{document}